\documentclass{article}

\PassOptionsToPackage{numbers, compress}{natbib}

\usepackage[preprint]{neurips_2024}

\usepackage{amsmath}
\usepackage{amsthm}
\usepackage[utf8]{inputenc} 
\usepackage[T1]{fontenc}    
\usepackage{hyperref}       
\usepackage{url}            
\usepackage{booktabs}       
\usepackage{amsfonts}       
\usepackage{nicefrac}       
\usepackage{microtype}      
\usepackage{xcolor}         

\usepackage[utf8]{inputenc}
\usepackage{multicol}
\usepackage[utf8]{inputenc}
\usepackage{graphicx}
\usepackage{titlesec}

\usepackage{geometry}
\usepackage{parskip}
\usepackage[official]{eurosym}
\usepackage{csquotes}

\usepackage{rotating}
\usepackage{lmodern}

\usepackage{wrapfig}
\usepackage[font=footnotesize,labelfont=bf]{caption}
\usepackage{mathtools}
\usepackage{indentfirst}

\usepackage{breakcites}

\usepackage{url}
\usepackage{graphicx}
\usepackage{bm}
\usepackage{enumitem}
\usepackage{cleveref}
\usepackage{float}

\definecolor{indigo}{RGB}{63, 81, 181}
\definecolor{red}{RGB}{210, 40, 95} 
\definecolor{pink}{RGB}{236, 64, 122}
\definecolor{green}{RGB}{46, 182, 125}
\definecolor{blue}{RGB}{66, 133, 244}
\definecolor{yellow}{RGB}{236, 178, 46}
\definecolor{anthracite}{RGB}{13, 13, 21}

\usepackage{amsmath,amsfonts,bm}

\def\eqref#1{equation~\ref{#1}}

\def\1{\bm{1}}

\def\vh{{\bm{h}}}

\def\vv{{\bm{v}}}

\def\vx{{\bm{x}}}

\def\vz{{\bm{z}}}

\DeclareMathAlphabet{\mathsfit}{\encodingdefault}{\sfdefault}{m}{sl}
\SetMathAlphabet{\mathsfit}{bold}{\encodingdefault}{\sfdefault}{bx}{n}

\DeclareMathOperator*{\argmax}{arg\,max}

\newcommand{\pred}{\bm{f}}
\newcommand{\predl}{\bm{f}_\ell}
\newcommand{\X}{\mathcal{X}}
\newcommand{\Y}{\mathcal{Y}}

\newcommand{\tr}{\mathsf{T}}

\renewcommand{\vx}{\bm{x}}

\renewcommand{\vh}{\bm{h}}
\renewcommand{\vv}{\bm{v}}
\renewcommand{\vz}{\bm{z}}
\newcommand{\btau}{\bm{\tau}}
\newcommand{\explainer}{\bm{\varphi}}

\title{Understanding Inhibition through Maximally Tense Images}

\begin{document}

\maketitle

\begin{center}

\textbf{Chris Hamblin}\textsuperscript{1,*} \footnote{ *email for correspondence: chrishamblin [at] fas [dot] harvard [dot] edu } \ \ \ 
\textbf{Srijani Saha}\textsuperscript{1} \ \ \ 
\textbf{Talia Konkle}\textsuperscript{1} \ \ \ 
\textbf{George Alvarez}\textsuperscript{1}

$^1$Harvard University, Department of Psychology 
\end{center}

\begin{abstract}
We address the functional role of \textit{feature inhibition} in vision models; that is, what are the mechanisms by which a neural network ensures images do \textit{not} express a given feature? Inhibition has received far less treatment in the literature than excitation, yet is critical for the construction of discriminative features. We observe that standard interpretability tools are not immediately suited to the inhibitory case, given the asymmetry introduced by the ReLU activation function. Given this, we propose inhibition be understood through a study of \textit{maximally tense images} (MTIs), i.e.  those images that excite and inhibit a given feature simultaneously. We show how MTIs can be studied with two novel visualization techniques; +/- attribution inversions, which split single images into excitatory and inhibitory components, and the attribution atlas, which provides a global visualization of the various ways images can excite/inhibit a feature. Finally, we explore the difficulties introduced by superposition, as such interfering induce the same attribution motif as maximally tense images.
\end{abstract}

\vspace{-.5mm}
\section{Introduction}
\vspace{-.5mm}

\begin{wrapfigure}[13]{rt}{0.3\textwidth}
  \centering 
 \includegraphics[width=0.28\textwidth]{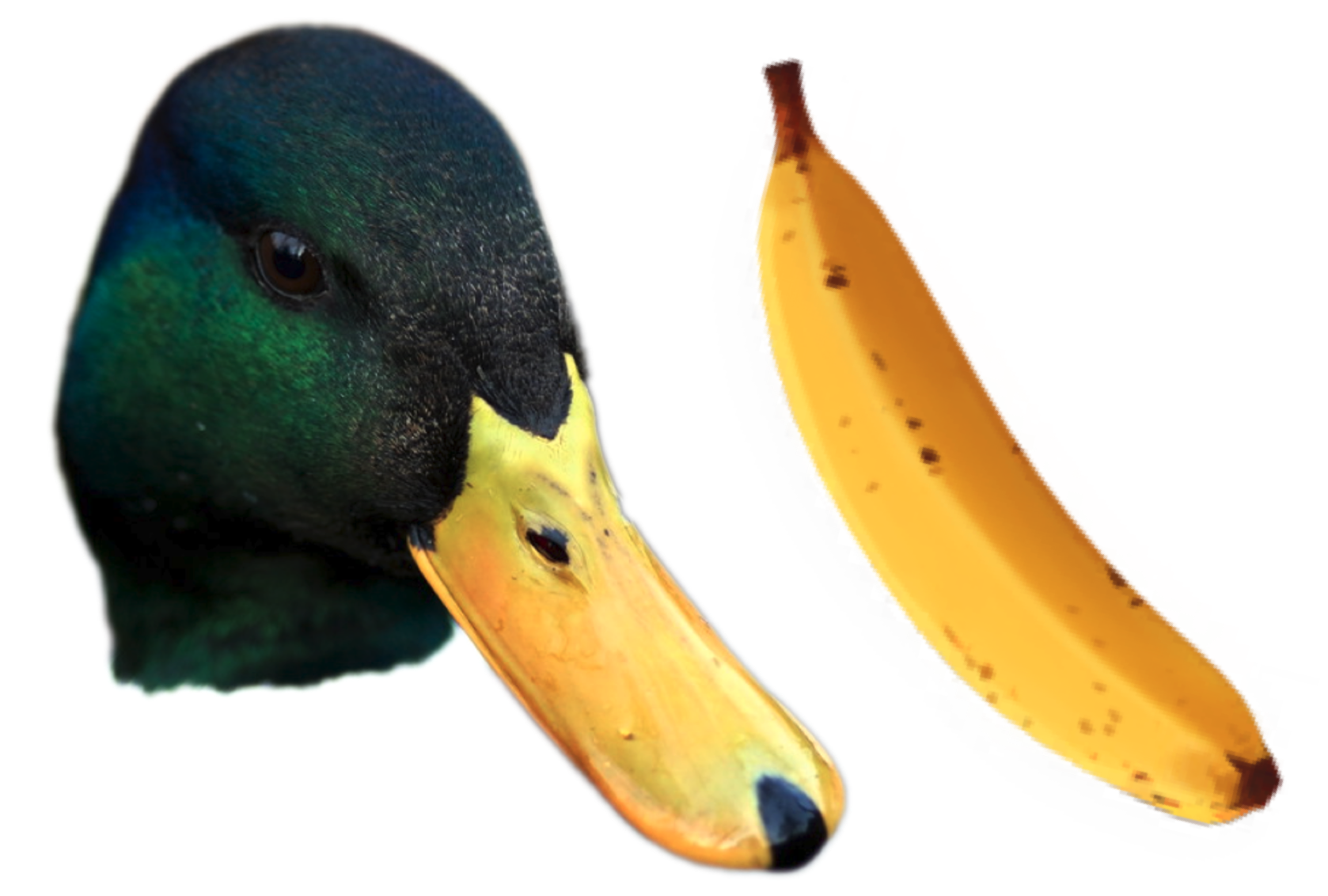}
  \caption{How might a network construct an accurate 'banana' feature, that doesn't activate for duckbills?
  }
  \label{fig:mask}
\end{wrapfigure} 
What makes an image \textit{not} activate a given feature in a neural network? This is the opposite of the question one typically asks, but it is important nonetheless; features are only useful if they are \textit{discriminative}, that is, if they activate in response to certain attributes of the input, but not others. A supposed 'banana' feature that activates for images of 'duck bills' isn't much of a 'banana' feature at all, and cannot be employed by the model as such. What, if any, are the mechanisms in a neural network that make features discriminative, that make duckbills \textit{not} bananas? If such mechanisms exist, how do we identify them? Do we need new tools, or is the current interpretability toolbox up to the task? \par
\paragraph{Maximally Exciting Images.} If we start by taking stock of this toolbox, we notice a common attribute of nearly every method is a reliance on \textit{maximally exciting images}, or MEIs\cite{klindt2023identifying}. In the general case, a feature can be thought of as a scalar-valued function of images, and an MEI is any input for which this function returns a large value. The simplest form of MEIs are the top-\(k\) activating images from a large dataset\cite{olah2017feature,borowski2020exemplary}, but more sophisticated interpretability techniques rely on them just them same. Feature visualization techniques synthesize MEIs with gradient ascent, in such a way that the optimized image expresses human-perceptible features, rather than adversarial ones\cite{olah2017feature,mahendran2015understanding,nguyen2015deep,tyka2016class,tsipras2018robustness,santurkar2019image,engstrom2019adversarial,nguyen2016multifaceted,mordvintsev2015inceptionism,wei2015understanding,nguyen2016synthesizing,nguyen2017plug}. Saliency map techniques return a heatmap over an image that highlights the most important regions of a given image for the expression of a given feature\cite{simonyan2013deep,bach2015pixel,baehrens2010explain,smilkov2017smoothgrad,sundararajan2017axiomatic,fel2021sobol,novello2022making,Fong_2017,zintgraf2017visualizing,petsiuk2018rise,fel2023don,ribeiro2016i,lundberg2017unified}. In effect, saliency maps reveal smaller, spatially localized MEIs that the user should identify with the feature. 'Concept'-based techniques specify the features in a model we should be studying in the first place\cite{TCAV,ghorbani2019towards,fel2023craft,fel2023holistic,voleurs-didee-nmf}, but these are typically paired with an assessment of MEIs, as features still need to be understood regardless of how they are identified in the model. Finally, where the above techniques characterize what features \textit{represent}, mechanistic approaches seek to explain how features are \textit{computed}\cite{circuitszoomin,elhage2021mathematical}. Mechanistic accounts of a model typically describe functions that operate on simpler/earlier features to compute complex ones. Even here, it is common practice to visualize the component features that comprise such functions with MEIs\cite{fel2023craft,circuit_curve,circuit_equivariance,carter2019activation}. \par

\begin{wrapfigure}[29]{rt}{0.5\textwidth}
  \centering
  \includegraphics[width=.49\textwidth]{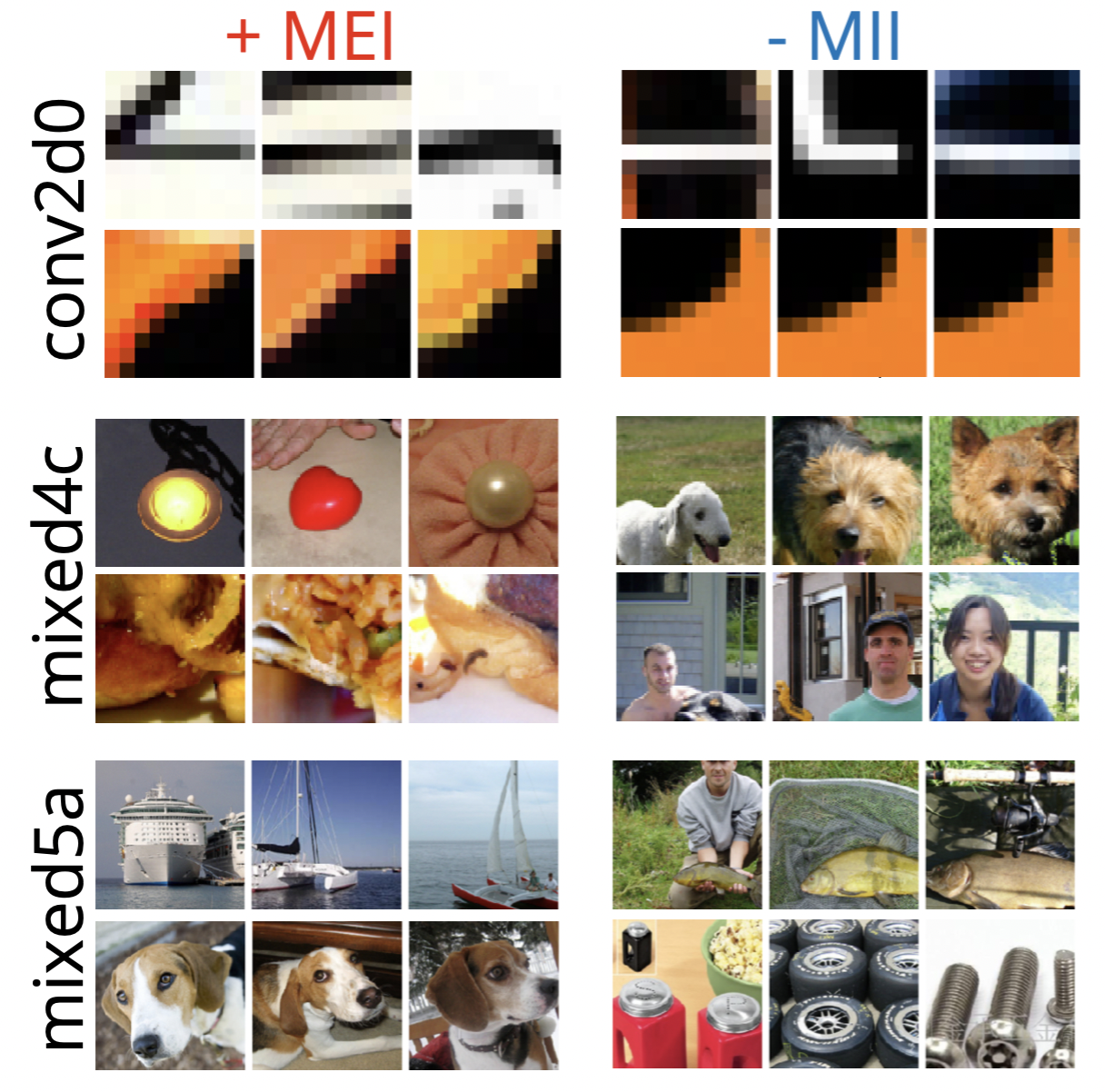}

  \caption{Imagenet \cite{imagenet_cvpr09} validation dataset example MEI and MII images for random features across several layers of InceptionV1 \cite{inception}. For each layer, the top row images correspond to MEIs/MIIs for a unit in that layer. The bottom row images correspond to a feature direction identified with k-means clustering. For both unit and k-means features, MEIs and their respective MIIs seem relatable in early layers, but arbitrarily paired in later layers.}\label{fig:MEI_MII_examples}
\end{wrapfigure}

\paragraph{Maximally Inhibiting Images.} Given MEIs are ubiquitous in our understanding of what features \textit{are}, a natural starting point for understanding what features \textit{aren't} is MIIs, or "maximally inhibiting images". Such images are sometimes considered in the literature; for example, when characterizing many units in a single model, each can be quickly conveyed through its set of MEIs and MIIs\cite{olah2017feature}. Additionally, human experiments on the interpretability of features often invoke MIIs in their design; a feature is considered 'interpretable' if humans can extrapolate from its set of MEIs and MIIs, correctly predicting whether new images belong to the MEI or MII set \cite{,borowski2020exemplary, zimmermann2021well, klindt2023identifying}. Mechanistic interpretability methods that consider inhibition also invoke MII; for example, by looking at the \textit{MEI}s for units connected by large negative weights\cite{circuitszoomin,circuit_curve,circuit_equivariance} or large negative attributions\cite{carter2019activation}.\par

MIIs may be a natural starting point for understanding what inhibits features, but they raise immediate concerns. First, consider the model architecture, specifically the ReLU activation function, $\text{ReLU}(\bm{x}) = \text{max}(\bm{x},0)$, which introduces an asymmetry between positive and negative activations. What is the use in knowing an image induces a large negative activation if this is precisely the information the ReLU function \textit{throws out}? Negative weights are not learned by the model \textit{so that} features return large negative values for certain inputs.
It's not clear what MIIs mean to the network, but there's also a second problem; in many cases MIIs are not meaningful to humans. This is an empirical point, by which we mean a feature's MEIs and MIIs often bear no visual relationship to each other, particularly for the high-level features represented in the later layers of the network. One might hope that MIIs have some property we can intuit as the 'opposite' of their respective MEIs, in which case the two image sets would constitute the poles of a meaningful axis. For example, consider the MEIs and MIIs for features in layer 'Conv2d0' of InceptionV1 \cite{inception}, shown in Figure \ref{fig:MEI_MII_examples}. In this first convolution layer the MIIs are approximately the MEIs but with their colors reversed; for example, orange above black becomes black above orange. However in late layers like 'mixed5a', where features have large receptive fields and rich semantics, the relationship between MEIs and MIIs seems arbitrary. We validate these intuitions with a human experiment, showing people can learn to predict MIIs from MEIs in the first layer, but not in later ones (appendix \ref{sec: human}).\par
To address these issues with MIIs, we present the following contributions:

\begin{itemize}
    \item We introduce analyses of \textit{maximally tense images} (MTIs), which excite and inhibit a target feature simultaneously. With regards to MTIs, negative weights play a meaningful functional role, as the feature would erroneously activate in response to its MTIs were it not for these weights.  
    \vspace{-.5mm}
    \item We present 2 novel feature visualization techniques that explain feature inhibition both locally and globally.
        \vspace{-.5mm}
    \item We explore how inhibitory weights facilitate superposition, and the difficulty this introduces for a mechanistic understanding of inhibition.
\end{itemize}

\paragraph{Notation.}
In what follows, we consider a neural network $\pred : \X \to \Y$, which transforms an input image, \(\vx \in \X\), through a sequence of $L$ hidden representations. Let $\predl : \X \to \mathcal{H}_\ell$ denote the function mapping the image to the \(\ell^{\text{th}}\) such  hidden representation, $\vh_{\ell} = (h_1, ..., h_{n_\ell})^\tr \in \mathcal{H}_\ell \subset \mathbb{R}^{n_\ell}$. In this work, a feature corresponds to a vector \(\vv \in \mathcal{H}_\ell\), and the function that computes the feature's 'activation' as $\pred_{\vv}(\vx) = \predl(\vx) \cdot \vv$.

\paragraph{Hardware \& Software.} The following experiments utilized 2 GeForce RTX 2080 GPUs. We used open-source software  Pytorch (BSD 3-Clause License) \cite{paszke2019pytorch}, and the \textit{faccent} (Apache License 2.0) \cite{faccent} feature visualization library.

\vspace{-.5mm}
\section{Completeness of gradient-based attribution}\label{sec:completeness}
\vspace{-.5mm}

Specifying MTIs will rely on computing 'attributions' for feature activations in an earlier layer of the model. In particular, we will leverage an empirical property of these attributions, that they behave \textit{additively}. Consider a fully linear model, \(y = \mathbf{w}\vx = w_{1}x_{1}+...+w_{n}x_{n}\). Observe that in this linear case \(w_{i} = \frac{\partial y}{\partial x_{i}}\), thus \(y = \nabla_{\vx}y\cdot \vx\). When \(y\) is a nonlinear function of \(\vx\), \(y = \nabla_{\vx}y\cdot \vx\) is a linear approximation, useful for many different applications, such as pruning \cite{braindamage,actgrad,snip} and saliency maps \cite{saliency,smilkov2017smoothgrad}. In our case, we want to understand the computation of feature \(\pred_{\vv}\) as some function of features computed in an earlier layer, so let's define a layer-to-feature attribution vector, \(\bm{S}_{l}\) and its sum, \(E_{l}\):

\begin{equation}
\bm{S}_{l}(\vx,\pred_{\vv}) := \nabla_{\vh_{l}} \, \bm{f}_{\bm{v}}(\vx)\odot\bm{h}_{l} \;\;\;\;\;\; E_{l}(\vx,\pred_{\vv}) := \nabla_{\vh_{l}} \, \bm{f}_{\bm{v}}(\vx)\cdot\bm{h}_{l}
\end{equation}

\begin{wrapfigure}[24]{r}{0.48\textwidth}
  \centering
  \includegraphics[width=.45\textwidth]{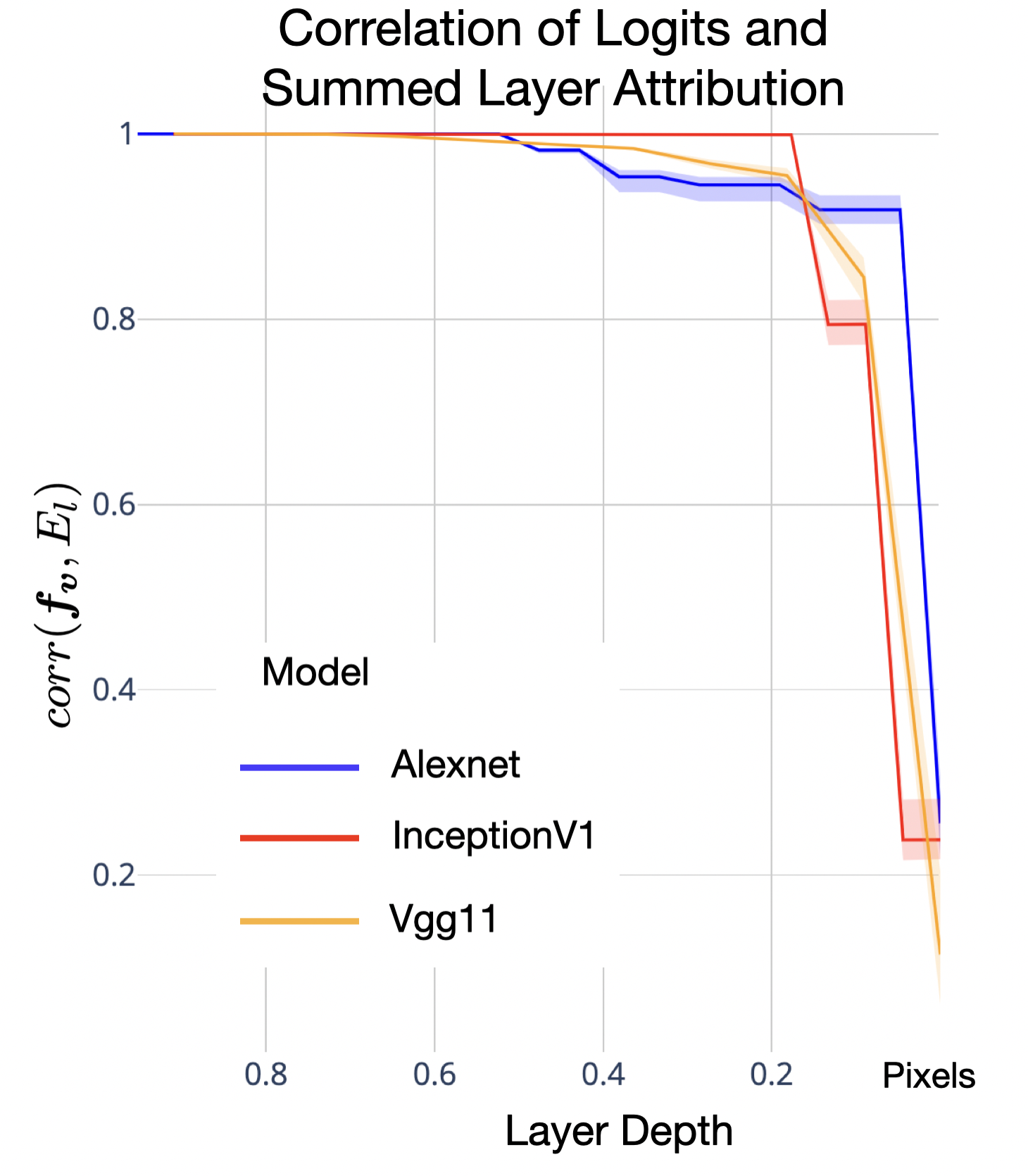}

  \caption{The correlation between logits, \(\pred_{\vv}\), and total attribution \(E_{l}\) measured across layers. \(\pred_{\vv} \approx E_{l}\) across all layers, except when measured through very early layers and pixels. 
  }\label{fig:energy_correlation}
\end{wrapfigure}

If \(\pred_{\vv} \approx E_{l}\), then \(\bm{S}_{l}\) can be used to select \textit{maximally tense images} (MTIs). MTIs are those images for which the attribution vector \(\bm{S}_{l}\) has both large positive and negative terms, indicating instances in which inhibition plays an important functional role, additively negating excitation. Previous work has called the property in which an attribution vector sums to \(\pred_{\vv}(\vx)\) 'completeness', and unfortunately has observed that \(E_{l}\) as defined does not satisfy completeness when attributing across entire image classification models, from a class probability to pixels\cite{axiomatic_attribution,shrikumar2017just,bach2015pixel}. However, \(E_{l}\) may still be complete when attributing between latent layers of the model, avoiding the gradient-flattening effects of the final softmax, as well as the non-linear relationship between pixel intensities and representations in later layers.\par

To test this, we computed \(E_{l}(\vx,\pred_{\vv})\) for 20 random logits and all Imagenet \cite{imagenet_cvpr09} validation set images across several models. Figure \ref{fig:energy_correlation} shows the average Pearson's correlation of \(\pred_{\vv}\) and \(E_{l}\) across logits. \(E_{l}\) is computed for all ReLU layers in the model and the pixel space, and Figure \ref{fig:energy_correlation} orders these measurements by layer depth. We find that \(E_{l}\) are close to ceiling in their correlation with logits when computed through most layers of the model. When measured through pixels however, the correlation deteriorates drastically. We find batch normalization layers also corrupt this relationship (appendix). \par 




\vspace{-.5mm}
\section{Attribution through a curve detector; a case-study}
\vspace{-.5mm}
\begin{figure}[htbp]
  \centering
  \includegraphics[width=\textwidth]{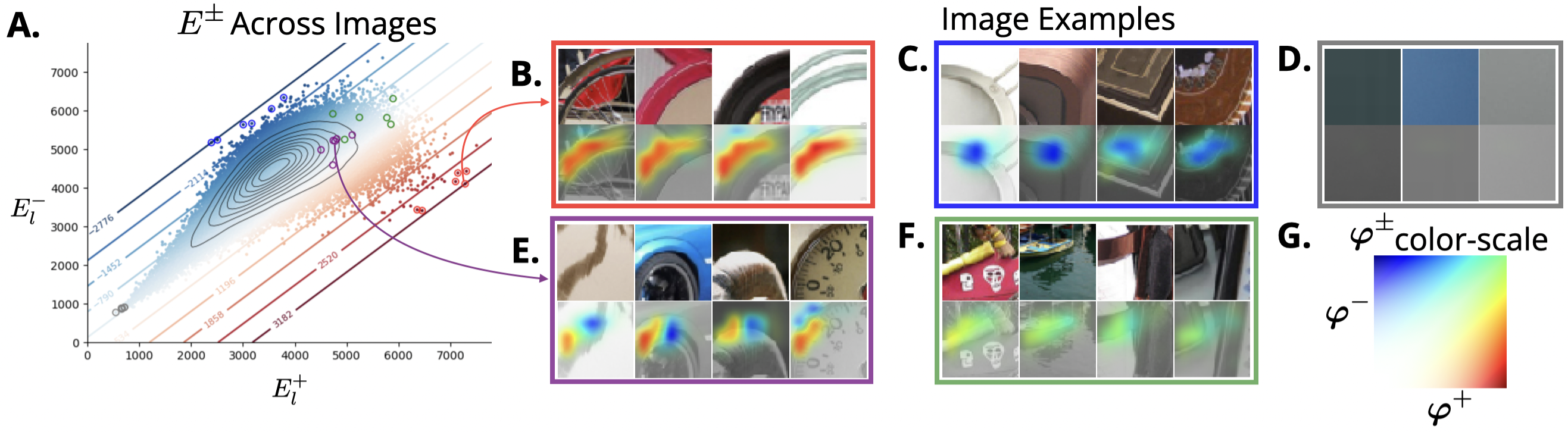}

  \caption{\textbf{A.} A scatterplot of \(E^{+}_{l}\) and \(E^{-}_{l}\) for a proposed 'curve detector' unit, across validation set images. Selected images visualized in \textbf{B.}-\textbf{F.} are circled in with the corresponding color. \textbf{B.} shows MEI examples, \textbf{C.} MIIs, and \textbf{D.} images with no attribution. \textbf{E.} shows images with positive and negative attributions in different spatial location, while \textbf{F.} shows images with positive and negative attribution in the channel dimension, at the same spatial location. \textbf{G.} The colorscale used for the \((\explainer^{+}_{l},\explainer^{-}_{l})\) cam maps, which spatialize the positive and negative attribution in a given image.}\label{fig:curve_full}
\end{figure}

We'll now explore several visualization techniques that leverage the completeness of the layer-to-feature attribution vector. Throughout, we'll apply these techniques to feature "mixed3b:379" in Inceptionv1 \cite{inception} -- a purported curve detector studied extensively in Cammarata et al. (2020)\citep{cammarata2020curve}. To begin, let's group \(\bm{S}_{l}\) into the sum of its positive and negative terms;

\begin{equation}
E^{+}_{l} := \sum_{i=1}^{n_{\ell}}{\text{ReLU}(\bm{S}_{l})_{i}}, \;\;\;\; E^{-}_{l} := \sum_{i=1}^{n_{\ell}}{\text{ReLU}(-\bm{S}_{l})}_{i}. 
\end{equation}

Figure \ref{fig:curve_full}.a shows the distribution of attributions \((E^{+}_{l},E^{-}_{l})\) for feature "mixed3b:379"  in the preceding layer. Each point represents the attributions for the central activation in the feature's activation map in response to the Imagenet validation set. The color of each point corresponds to the activation value (pre-ReLU), and the diagonal lines are contours lines for \(E_{l}\) using the same color scale. Because \(\pred_{\vv} \approx E_{l}\), the coloring of the data points and the contour lines are well-aligned. \par
Some data points have been selected from Figure \ref{fig:curve_full}.a, with the corresponding images shown in Figures \ref{fig:curve_full}.b-f (cropped at the receptive field). The first and second of these selections correspond to the MEIs (b.) and MIIs (c.), which show excitatory and inhibitory curves at opposing orientations, as noted in Cammarata et. al. (2020)\citep{cammarata2020curve}. A saliency map, similar to GradCAM \cite{GradCAM}, accompanies each image, which serves to spatially localize our attributions. Observe that when layer \(l\) is convolutional, attribution vector \(\bm{S}_{l}(\vx,\pred_{\vv})\) is actually a tensor in \(\mathbb{R}^{C \times H \times W}\). Spatial maps analogous to \(E_{l}^{\pm}\) can be computed by summing over only the channel dimension,\footnote{\(E^{\pm}\) refers to the tuple \((E^{+},E^{-})\).
Well use the superscript \(\pm\) analogously for related variables.}

\begin{equation}
\explainer^{\pm}(\pred_{\vv}, \vx) := \sum_{c=1}^{C}{\text{ReLU}(\pm\bm{S}_{l,c}(\pred_{\vv}, \vx)}).
\end{equation}

Such maps can be viewed as a heatmap over the image by upsampling them to the input image dimensions, as with GradCAM. We can visualize the negative and positive maps simultaneously by using a special color scale, which we show in Figure \ref{fig:curve_full}.g. With this coloring excitatory image regions appear red, inhibitory regions appear blue, and regions that have positive and negative terms in the channel dimension appear green. \par
 These maps are normalized \textit{across} images, making it possible these attribution maps to highlight no regions of the image. For example, Figure \ref{fig:curve_full}.d shows those inputs that have near zero \(E^{+}_{l}\) and \(E^{-}_{l}\). The cropped images show homogeneous color patches that neither excite nor inhibit the curve detector, and the attribution maps highlight nothing in these crops.\par
The images of most interest for our purposes are those which yield large values for both \(E^{+}_{l}\) and \(E^{-}_{l}\). Such images are depicted in Figure \ref{fig:curve_full}.e and .f, which each illustrate distinct cases. In the first case, inhibition and excitation to the curve detector come from distinct regions of the image. These images were selected from the set for which \(\explainer^{+}(\vx) - \explainer^{-}(\vx)\) contains both values \(< P_{1}\) and \(> P_{99}\). These images are easy to understand; they contain an excitatory curve at a consistent position/orientation, but also an inhibitory curve in a different location. The images in Figure \ref{fig:curve_full}.f depict a different case, in which simultaneous excitation/inhibition happens in the channel dimension at a single spatial position. These images were selected from the set for which both \(\explainer^{+}(\vx)\) and \(\explainer^{-}(\vx)\) contain a value \( > P_{99}\) at the same spatial position. These images are harder to interpret, as the heatmap directs our attention to the same region to explain what excites and inhibits the curve detector. How should we understanding this channel-wise inhibition and excitation of a feature in the general case? \par

\begin{figure}[tbp]
  \centering
  \includegraphics[width=\textwidth]{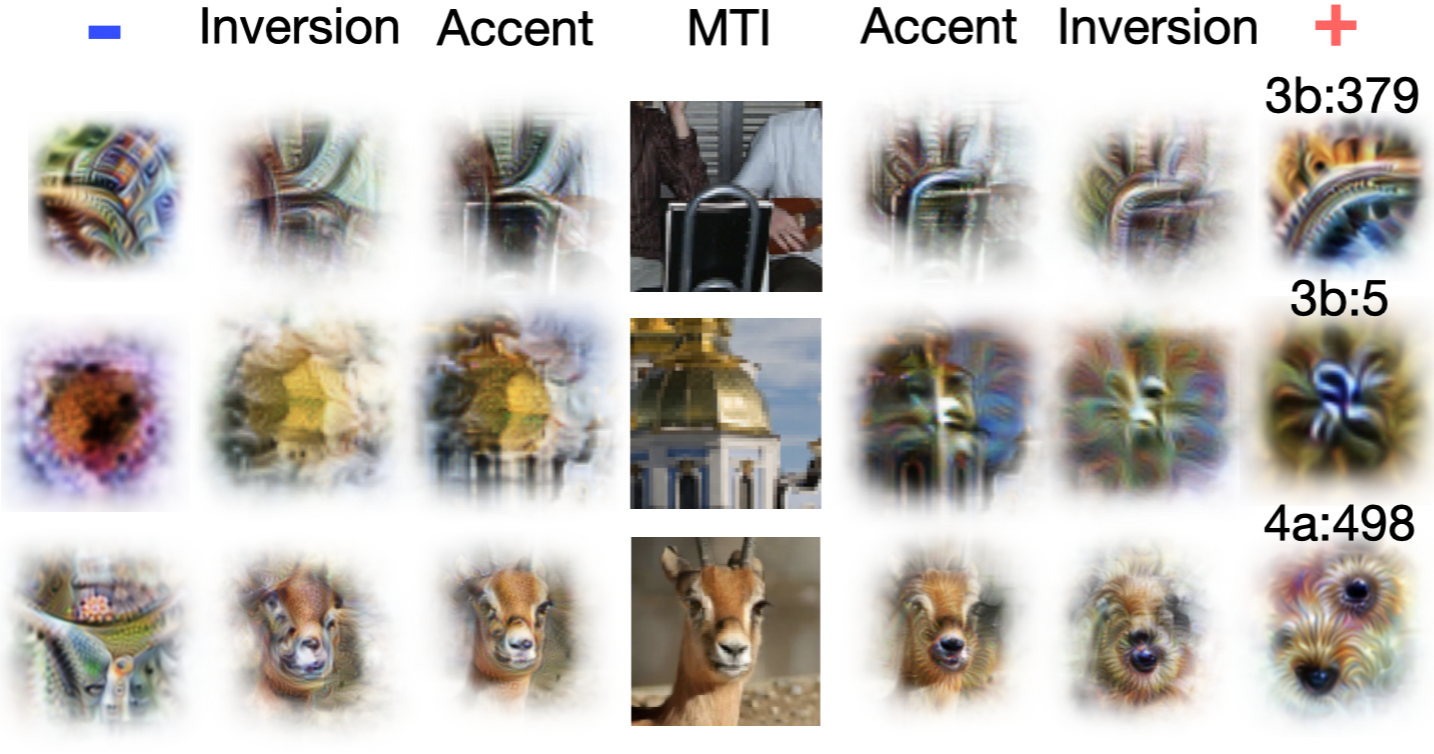}

  \caption{MTIs for 3 units, with their \(\pm\) attribution accentuations, inversions, and standard feature visualizations}\label{fig:MTI faccent}
\end{figure}

\vspace{-.5mm}
\section{Accentuating and Inverting MTI Attributions}\label{sec: attr inversion}
\vspace{-.5mm}
For a given image, we want to exaggerate \(\bm{S}_{l}^{\pm}\), which we'll do by treating these attributions themselves as the feature activations to be maximized. This is similar to a technique utilized in \citet{cammarata2020curve}, but isolating positive and negative attributions separately. Additionally, rather than maximize the dot product with the attribution vector, we'll scale the dot product by the cosine similarity as in \citet{carter2019activation}, which encourages the optimized image to point in the same direction as \(\bm{S}_{l}^{\pm}\) (appendix \ref{sec:dot*cosine}). As is typical with feature visualization, we can optimize the image with a parameterization \(\mathcal{P}(\vx)\) \cite{mahendran2015understanding,olah2017feature,mordvintsev2018differentiable} and under a set of transformations \cite{mordvintsev2015inceptionism} $\btau \sim \mathcal{T}$, giving us the optimization;

\begin{equation}\label{eq: attribution viz}
\vz^{\ast} = \argmax_{\vz} \mathcal{L}(\btau \circ \mathcal{P}^{-1}(\vz);\bm{S}_{l}) 
~~ with ~~ \mathcal{L}(\vx;\bm{S}_{l}) := \frac{(\pred_{l}(\vx) \cdot \bm{S}_{l})^{p+1}}{(||\pred_{l}(\vx)||\cdot||\bm{S}_{l}||)^{p}}
\end{equation}

We can view the image that results from this as \(\vx^{\ast} = \mathcal{P}^{-1}(\vz^{\ast})\), and optionally seed from noise (inversion) or the natural image (accentuation) we used to calculate \(\bm{S}_{l}\). In Figure \ref{fig:MTI faccent} we show accentuations and inversions for MTIs across 3 InceptionV1 units, including our curve detector from earlier. Each MTI is among the top 10 images with the largest norm, \(||\bm{S}_{l}(\vx)||_{1}\), under the constraint that \(.5\sigma({\pred_{\vv}(\bm{X})})<\pred_{\vv}(\vx)<0\), which ensures excitation and inhibition are balanced. We use the fourier phase space image parameterization from MACO\cite{fel2023maco}, as well as their opacity masking technique, which integrates the pixel gradients over the optimization steps and sets this to the alpha channel of the image. We use randomly located crops that cover .9-.99 of the entire image, as well as uniform and gaussian noise, as our set of transformations.\par

\begin{figure}[tbp]
  \centering
  \includegraphics[width=\textwidth]{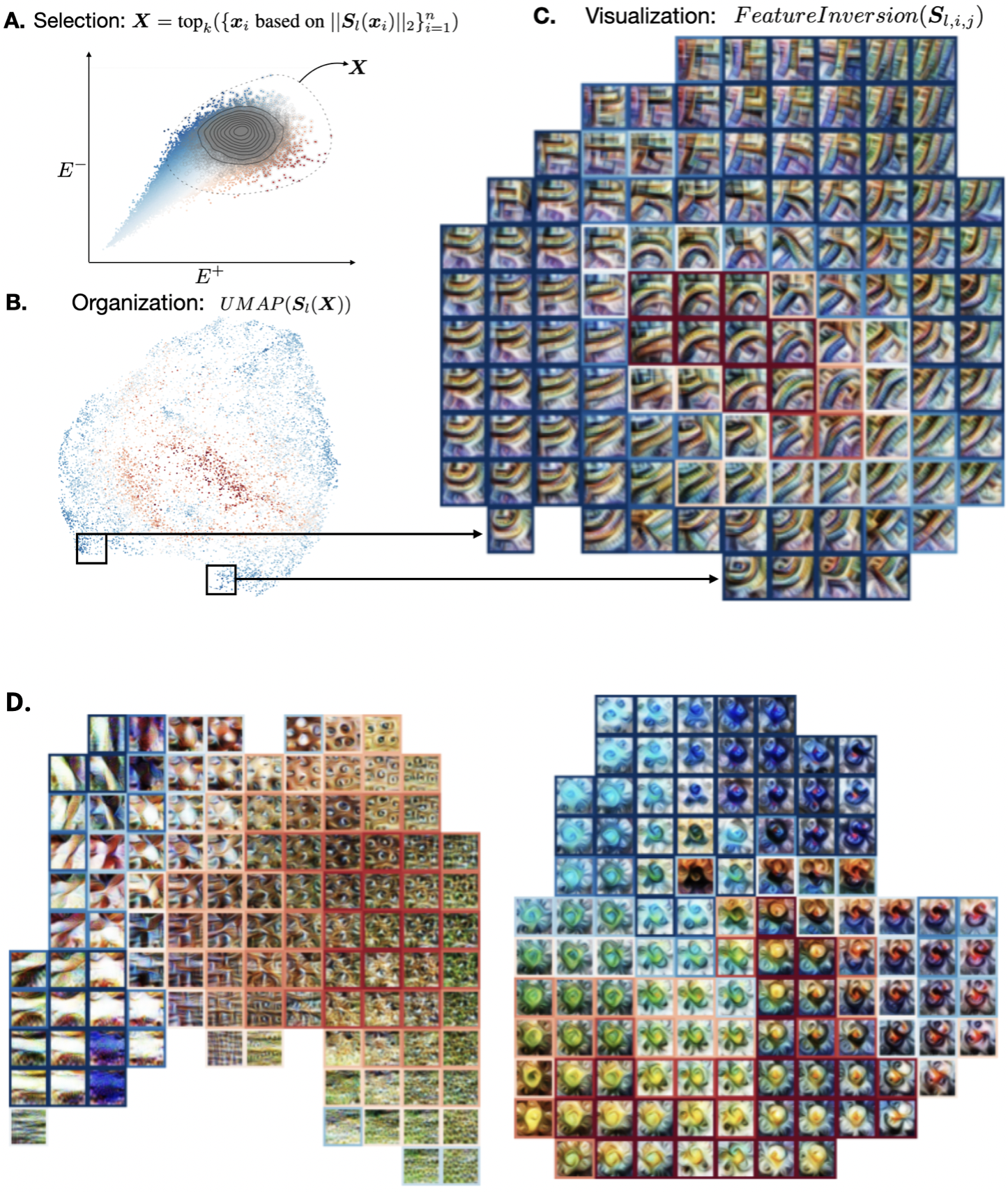}

  \caption{\textbf{A.} Selection of large attribution examples from the dataset shown over the plot of \(E^{\pm}\) values. Contour lines show the density of selected points in the plot. \textbf{B.} Organize the attribution vectors of these selected points using UMAP. \textbf{C.} Average the attributions within local regions of the UMAP, then perform feature attribution inversion (equation \ref{eq: attribution viz})}\label{fig:atlas_demo}
\end{figure}

\vspace{-.5mm}
\section{Feature Attribution Atlas}\label{sec:attr atlas}
\vspace{-.5mm}

Feature inversion and accentuation can help us understand how an individual image can excite and inhibit a feature, but what if we want a global view? That is, can we generate a visualization that depicts the relationships between all the various ways different images excite/inhibit a feature? Here we will adopt techniques from the \textit{activation atlas}\cite{carter2019activation}, which combines \textit{UMAP}\cite{umap} and feature visualization to map the space of activations of whole neural network layers. Here we propose the \textit{Feature Attribution Atlas}, which conditions this technique on single features, mapping the space of feature attributions in a layer. Generating the atlas is a three step process. 

\paragraph{Image Selection:} The atlas should be a function of those images relevant to the target feature, which we can be determined by the its attributions. For a large set of images,\(\mathcal{D} = {\bm{x}}_{i}^{n}\), we compute \(\bm{S}_{l}(\mathcal{D}\), then select a subset of images \(\bm{X}\) to construct our atlas. In this demonstration we use 100,000 ImageNet training images as \(\mathcal{D}\) and select \(\bm{X}\) as the top 10,000 with the largest \(L^2\) attribution norm, \(||S_{l}(\bm{x})||_{2}\). This selection criteria allows our atlas to convey those images which excite and/or inhibit \(\pred_{\vv}\), but not those images which are strictly orthogonal. We find using the \(L^{2}\) norm yields more diverse atlases than using \(L^{1}\), see section \ref{sec:uniqueness} for details on why this might be the case.



\paragraph{Attribution UMAP:} Next we organize images by their attribution vector using UMAP; \(map_{\bm{S}} = umap(\bm{S}_{l}(\bm{X}))\). This map conveys the various 'reasons why' images excite/inhibit \(\pred_{\vv}\), as points close in  \(map_{\bm{S}}\) will correspond to images with a similar attribution vector. We color points in the map by the correspond activation of \(\pred_{\vv}\), to get a sense for where inhibitory and excitatory images land in the map (Figure \ref{fig:atlas_demo}.b). 

\paragraph{Feature Inversion:} Its difficult to visualize all \~ 10,000 images represented in \(map_{\bm{S}}\) directly, so we perform a coarse-graining operation. We overlay \(map_{\bm{S}}\) with an \(n \times n\) grid, then average all the attribution vectors within each grid cell, associating a new average attribution \(S_{l,j,i}\) with each position \((i,j)\) in the grid. We can then generate a visual icon for each position in the map using our attribution inversion visualization technique (equation \ref{eq: attribution viz}). Where in section \ref{sec: attr inversion} we visualized \(\bm{S}_{l}^{+}\) and \(\bm{S}_{l}^{-}\) independently to isolate the inhibitory and excitatory influences from an MTI, in this application we will optimize towards \(|\bm{S}_{l,i,j}|\), which will allow positive and negative influences to be expressed in a single icon when in a 'tense' region of the map. Finally, we pass these icon images back through the network and compute \(\pred_{\vv}\), then color the icon border with its activation value, and position the icons in the corresponding grid location, yielding the attribution atlas (Figure \ref{fig:atlas_demo}.c).

Viewing a feature's attribution atlas can provide much insight over dataset examples and feature visualizations alone. We find it particularly useful for understanding the role of negative weights into a feature. For example, mixed3b:9, appears to be inhibited by low frequencies, given its negative feature visualization, but as discussed earlier, its not sensible to conclude these weights are present \textit{so that} the feature returns large negative activations to low frequency inputs. Rather we can understand the functional role of inhibition through the 'tense' regions of the attribution atlas (Figure \ref{fig:atlas_demo}.c), where  inhibition and excitation meet. Inhibition ensure the feature returns only moderate activations in response to such inputs. 

\vspace{-.5mm}
\section{Inhibition from Superposition}
\vspace{-.5mm}

Thus far we have presented a framework for understanding the role of inhibition in the construction of features. The lynch-pin of our approach is the attribution in intermediate layer \(l\); inhibition plays a functionally relevant role in how \(\pred_{\vv}\) is computed for input \(\vx\) when \(E_{l}^{\pm}(\vx,\pred_{\vv})\) are both large. 
Suppose though, that we found an image \(\vx\) expressing only features independent of \(\pred_{\vv}\), that nonetheless induced a large \(E^{\pm}(\vx,\pred_{\vv})\). This would present a problem, as such an image expresses no features relevant to the computation of \(\pred_{\vv}\), and scrutinizing such an image would only mislead us. However, recent work on toy models of superposition\cite{superposition} suggests that such attributions are possible. Specifically, when features are sparse the model may represent more features than it has dimensions, Such that independent features are represented with \textit{interference} (non-zero dot product) in a latent layer. 
It was hypothesized that reading out from features in superposition could necessitate inhibition negating the excitation caused by interference, in which case we would observe large \(E_{l}^{\pm}(\vx,\pred_{\vv})\) even when all features expressed by \(\vx\) are independent of \(\pred_{\vv}\). \par

\begin{wrapfigure}[22]{R}{0.42\textwidth}
  \centering
  \includegraphics[width=.4\textwidth]{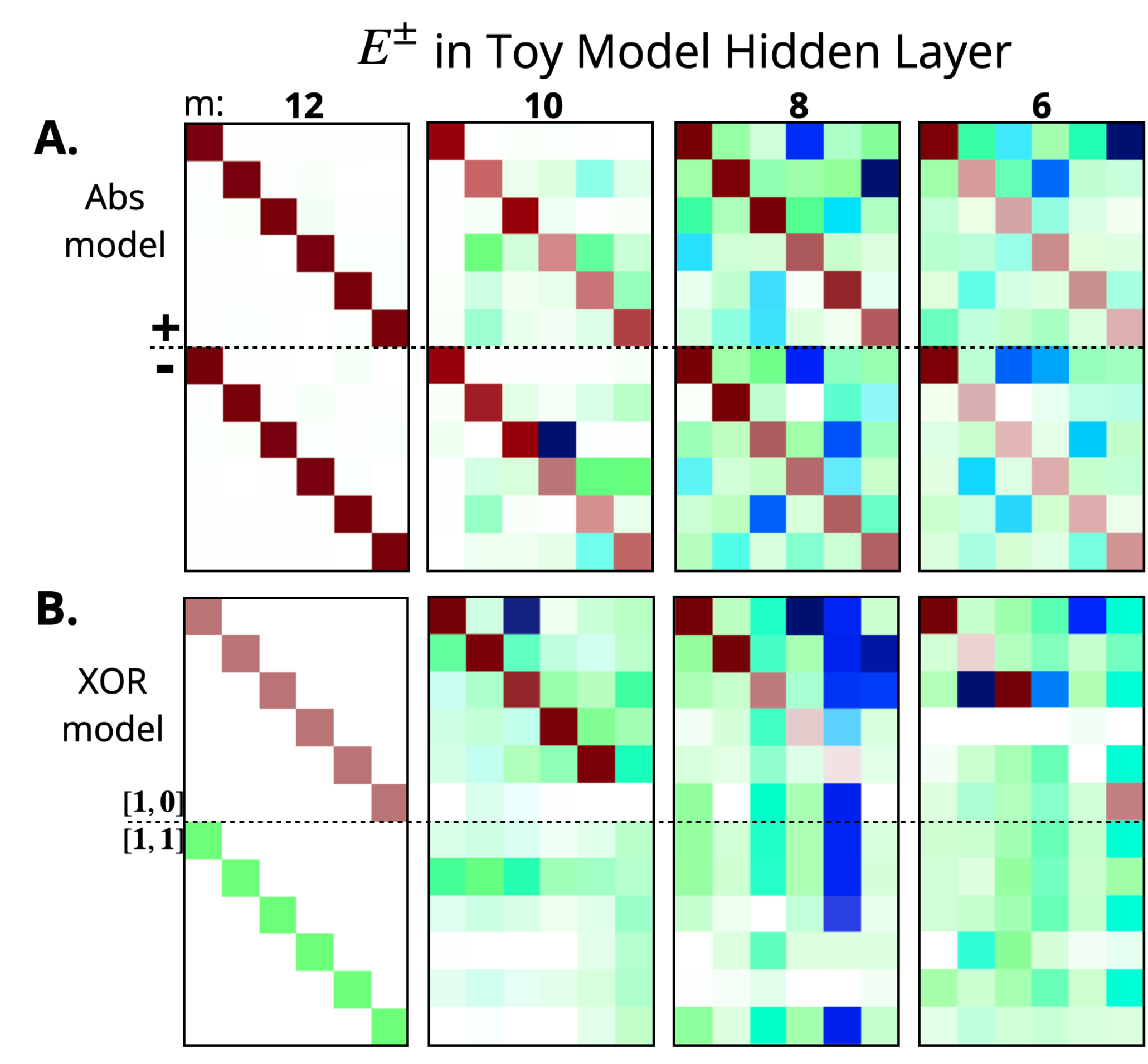}

  \caption{Toy models compute the (\textbf{A.}) absolute value function and (\textbf{B.}) XOR function on 6 independent input features. 
  Attributions are normalized within each column. For a color-scale legend, see Figure \ref{fig:curve_full}.b}\label{fig:toy_model_energies}
\end{wrapfigure}

\paragraph{Superposition in toy models.} Following Elhage et al. (2022)\cite{superposition}, let's test that these deceptive attributions are possible with a toy model computing the absolute value function. 
It is a minimal non-linear function that is easy to compute with a small ReLU model, as \(\text{abs}(x) = \text{ReLU}(x)+\text{ReLU}(-x)\). 
For multivariate input \(\vx \in \mathbb{R}^{n}\), the network requires \(m=2n\) hidden neurons to compute \(\text{abs}(\vx)\) exactly.
In the original work, the authors find the network will optimize for very different weight motifs when \(m<2n\) and \(\vx\) is sampled sparsely, such that any given \(x_{i}\) usually takes the value 0. In our replication of these experiments, we train 4 models to compute \(\text{abs}(\vx)\) for \(\vx \in \mathbb{R}^{6}\), with hidden dimensions \(m=12,10,8,6\). As \(m\) decreases, the 6 input features should be represented with more interference in the hidden layer. \par

For each trained model we pass 12 inputs, one-hot vectors for each of the features and their negatives, i.e the basis vectors \(e_{i}\) and \(-e_{i}\). For each of these inputs we compute the attribution to each output feature, \(f_{i}\), which we've trained to compute \(f_{i}(\vx) = \text{abs}(x_{i})\). These attribution matrices are shown in Figure \ref{fig:toy_model_energies}; element \((i,j)\) shows \(E_{l}^{\pm}(e_{i},f_{j})\) in the hidden layer, colored according the legend in Figure \ref{fig:curve_full}.b. The matrix in the first column corresponds to a disentangled model that can compute absolute value exactly without feature interference. The matrix shows + attribution along the diagonal only, and 0 attribution everywhere else, meaning input \(e_{i}\) excites \(f_{i}\), but \(f_{j}\) is insensitive to \(e_{i}\) whenever \(i \neq j\), as it should be. Subsequent columns show attribution in models with ever tighter bottlenecks, inducing more interference, and subsequently more off-diagonal attributions. While the diagonal in these matrices is the only place where \(E^{+} > E^{-}\), resulting in the correct output feature activating and performant models (appendix \ref{fig:toy model loss}), there is still a lot of attribution in the off-diagonals. All of this off-diagonal inhibition and excitation is induced by independent features irrelevant to the computation of \(f_{j}\).\par

It's clear that superposition can induce the +/- attribution motif, however we know that inhibition is not only relevant for compression, as there are functions that provably cannot be approximated without negative weights, such as the xor function\cite{inhibition2023wang}. Figure \ref{fig:toy_model_energies}.b shows a similar experiment on a toy model computing the XOR function over 6 independent \textit{pairs} of features, flattened into an input in \(\mathbb{R}^{12}\). Instead of testing this model on inputs \(\pm e_{i}\), we pass the inputs \([1,0]_{i}\) and \([1,1]_{i}\) --  one element relevant to \(f_{i}\) is on, or both -- for which the model should return \(e_{i}\) and \(\mathbf{0}\) respectively. In this case the disentangled model shows simultaneous inhibition and excitation to \(f_{i}\) from input \([1,1]_{i}\); excitement from the latent 'or' feature and inhibition from the latent 'and' feature. This is the sort of functionally relevant inhibition we hoped to identify with MTIs, where inhibition cancels the effect of excitation, leading to no activation of the downstream feature. However, when this XOR model is implemented with a latent bottleneck, we see large off diagonal attributions just as before. \par

\paragraph{Superposition at scale}\label{sec:uniqueness}

How can we distinguish between instances of functionally relevant inhibition -- that is necessary for computing \(\pred_{\vv}(\vx)\) even in a disentangled model -- and instances of inhibition caused by superposition? In toy models its possible to make this distinction, because we know a priori how the features can be computed from the inputs. 
However, in full-scale object recognition models we don't know the ground-truth for how any feature is computed, that is precisely what we are trying to uncover empirically. 
One indication that a large attribution norm might be the result of superposition interference is it should affect multiple features. In the limiting case, a maximally superimposed feature in layer \(l\) points in the diagonal direction, and \textit{every feature} in layer \(l+1\) is excited and inhibited (presuming it has both positive and negative incoming weights) whenever this diagonal feature is expressed. Additionally, it has been theorized that superposition is less likely to occur in the early layers of an object recognition model, because the features represented in such layers, such as edges, are not sparse\cite{superposition}. 
Taking these two observations together, we should expect that in deeper layers of the network, many features show a high attribution norm to the very same images, as these images express highly interfering features. To test this, we consider a uniqueness metric across a sample of features in a layer. Suppose we have \(n\) features, and a set of \(m\) images, \(X_{i}\), is identified with each feature, \(\bm{f}_{i}\),  by some selection process. We can define a \textit{uniqueness} measure for this process as;

\begin{equation}
U = \frac{|\bigcup_{i=1}^{n} X_i|}{nm}
\end{equation}

\begin{wrapfigure}[25]{R}{0.48\textwidth}
  \centering
  \includegraphics[width=.45\textwidth]{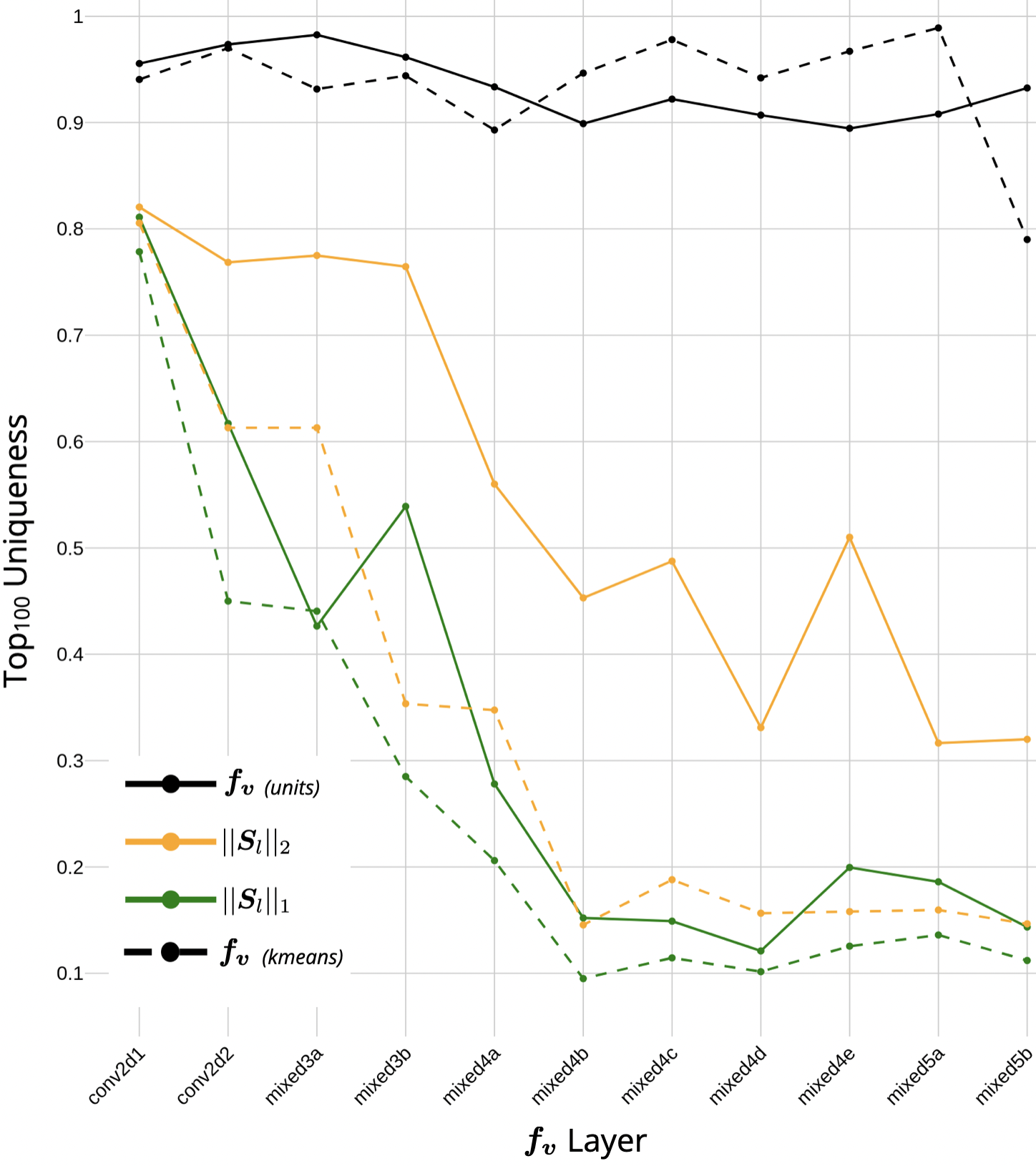}

  \caption{While features may show large activations for different images (black line), in deeper layers of the model they show large attributions to the same images, (green and orange lines). This is true of unit features and k-means features (dotted lines)}\label{fig:uniqueness}
\end{wrapfigure}

Observe that U is bounded above by \(1\), when every image selected is unique, and bounded below by \(\frac{1}{n}\), when \(\forall i, j \in \{1, \ldots, n\}, \, X_i = X_j\).\par
In Figure \ref{fig:uniqueness} we see how different image selection processes differ in this uniqueness measure. The black solid line shows the uniqueness of \textit{MEI}s across layers of InceptionV1 for 20 random units per layer, for which  \(U\) is near ceiling across all layers.
However, when those images with a large attribution norm (MTIs) in the preceding layer are identified with a unit, rather than those with a large activation, uniqueness decreases substantially in the deeper layers, as we predicted under the superposition hypothesis. Using the \(L^{2}\) norm of the attribution vector (orange line) rather than the \(L^{1}\) (green line), mitigates this somewhat, as the \(L^2\) norm grows less quickly than the \(L^1\) in off-basis directions\cite{l2l1_xu2021analysis,l2l1_yin2014ratio}, making \(L^2\) potentially less likely to select for superimposed features in the attribution layer. 
See the appendix \ref{sec:appendix uniqueness}  for examples of those images that yield high attributions across features.\par

\vspace{-.5mm}
\section{Limitations \& Conclusion}
\vspace{-.5mm}

Inhibition and excitation in ReLU neural networks do not function symmetrically, but currently the few interpretability tools that target inhibition do not account for this. We introduce several new techniques to the toolbox that respect this asymmetry, by conditioning our understanding of inhibition on excitation through the analysis of \textit{maximally tense images}. Our novel visualization techniques reveal how inhibitory connections prevent erroneous activation in response to MTIs, by isolating the inhibitory and excitatory attributes simultaneously present in such images. However, we also show that superposition currently introduces a major obstacle for these kinds of analyses, as networks use negative weights to facilitate a compression algorithm, representing more features than units. Given this, a 'clean' understanding of inhibition in deep layers of vision models will likely require the development of techniques for 'monosemantic disentanglement', as is being pursued for large language models\cite{bricken2023monosemanticity,cunningham2023sparse}. Additionally, we note that the inhibitory mechanisms described in this work need not be the only ones through which features are suppressed. For example, the final softmax layer of a classifier constitutes a different mechanism by which one 'class' feature inhibits another. Other model architectures could invoke similar suppressive mechanisms throughout, such as those using Top-k \cite{topk_act_1,topk_act_2} or SoLU \cite{solu} activation functions. In conclusion, we hope this work prompts more exploration into the extensive inhibitory mechanisms latent in vision models, and their critical role in feature construction.

\clearpage

\bibliography{references}
\bibliographystyle{iclr2024_conference}

\appendix
\newpage

\section*{Appendix}

\renewcommand{\thesection}{\Alph{section}}


\section{Code}

 Code for this project is available at \href{https://github.com/chrishamblin7/feature_inhibition}{this github repository}.

\section{Human Experiment}\label{sec: human}

\begin{figure}[H]
   \centering
   \includegraphics[width=\textwidth]{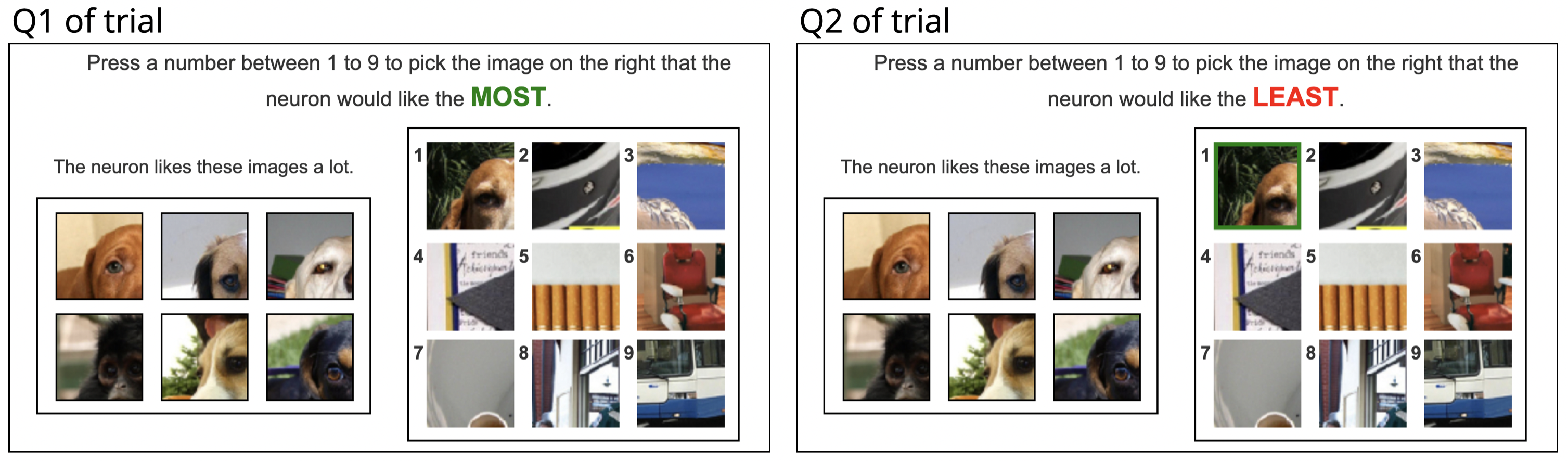}
   \caption{An example trial of our human experiment, in which participants must extrapolate from a set of MEIs to both a new MEI and an MII. In Q2 of the trial, the green outline for choice 1 indicates that it was selected as the new MEI. Participants could not choose the same image as the MEI and MII.}
\end{figure}\label{fig:trial_human_exp}

\begin{wrapfigure}{R}{0.46\textwidth}
   \centering
   \includegraphics[width=.45\textwidth]{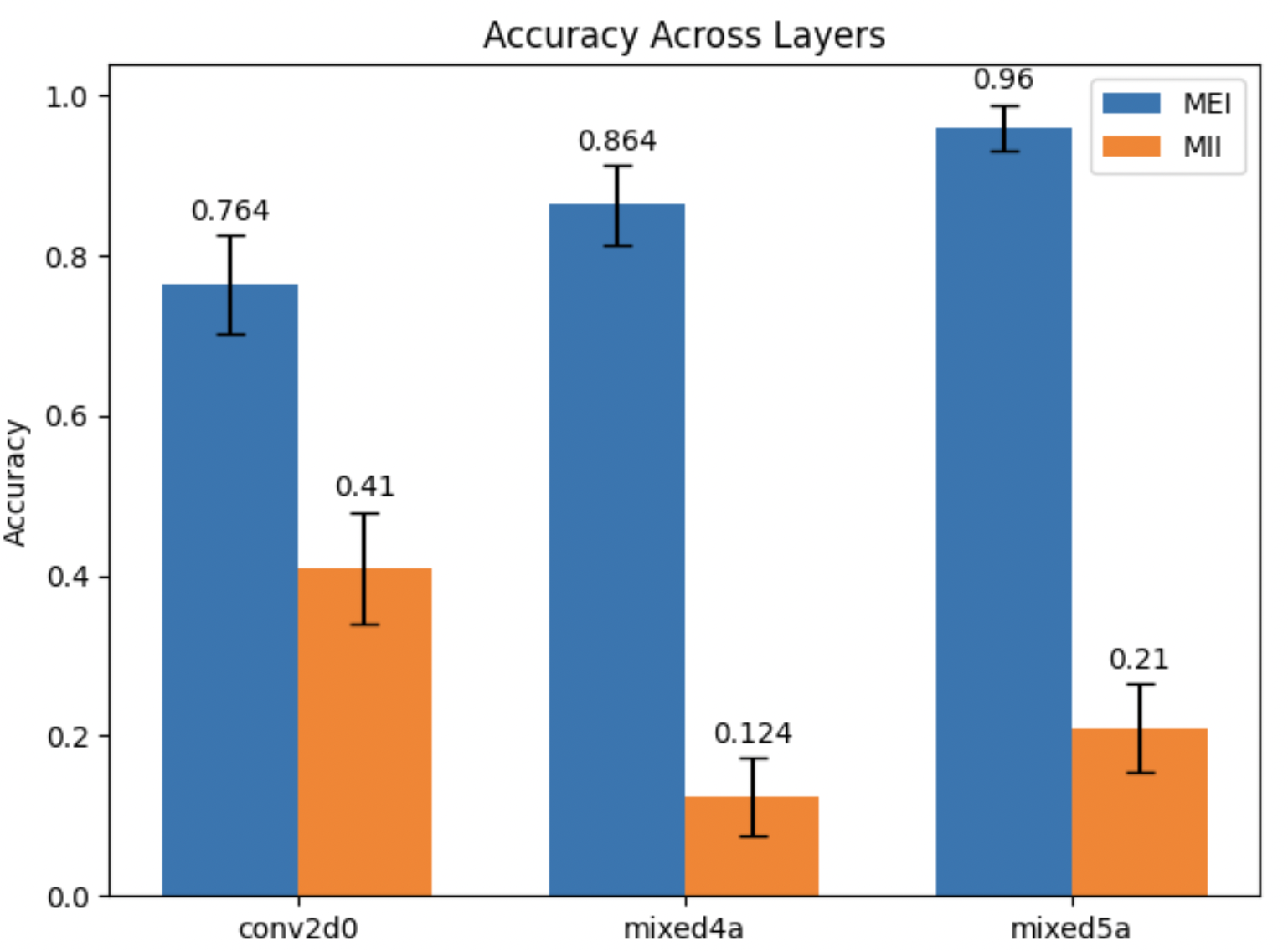}
   \caption{Human accuracy in predicting MEI/MIIs across model layers.}
\end{wrapfigure}\label{fig:human_exp_accuracy}

Here we explain a human experiment in which we endeavor to identify if/when people can relate the MEIs and MIIs of a feature. Given visual inspection of such exciting/inhibiting images, like those shown in figure \ref{fig:MEI_MII_examples}, we hypothesize that early on in the model humans will be able to conceive of a feature's 'opposite' given example MEIs, but wont be able to do so in later layers. To test this we conducted an experiment in which participants are shown a set of MEIs for a feature, then must use this information to identify the feature's MII from another set. \par
In this experiment we test 15 features per an early, mid, and late layer of InceptionV1 -- Conv2d0, Mixed4a, and Mixed5a. We use the centroids of k-means clusters as our feature directions, following the observation in \citet{klindt2023identifying} that shows such k-means features are particularly interpretable for humans in this experimental paradigm. In particular, we first specify 1000 clusters per layer using the SKlearn k-means clustering algorithm with a cosine distance metric. The clustering was applied to each layers' hidden vectors in response to the ImageNet validation set, with each image sampled at a random position in the activation map for each hidden vector. We chose 15 random features from these centroids per layer for use in this experiment. We defined the activation for such a feature in response to an image as the cos*dot (see section \ref{sec:dot*cosine}) of the hidden vector for the image in the corresponding pre-relu layer, computed in the channel dimension. We use \(p=2\) cosine power, which ensures cosine and dot product terms do not cancel out negatives when multiplied together. \par
For each feature identified this way we construct a trial of the experiment, which consists of 15  image (crops). First, we compute activations for the feature in response to the ImageNet validation set. The image that induces the largest activation we specify as the trial's 'target MEI', which is cropped to the effective receptive field \cite{recep_field_2} that induced the large activation. Similarly the smallest activation is specified as the 'target MII'. The 2nd-7th largest activations are specified as the 'example MEIs'. Finally 7 'distractor' image crops are selected, which each satify the simultaneous constraints of yielding only modest activation for the feature -- within 2 standard deviations of the mean --  and having large over all activation -- the hidden vector for the crop in the feature's layer has an \(L^{1}\)-norm in the 90th percentile. This second constraint ensures distractors are not different in kind from the target MEI and MII, conveying salient objects rather than awkward crops or background elements. Across trials of features in a given layer, we apply the additional constraints that crops cannot be repeated for any MII or distractor, and crops cannot come from different locations of the same image. \par
In a given trial of the experiment, the participant is shown the set of 6 'example MEIs' for the feature on the left. They are then shown a set of 9 images on the right; the 7 'distractors', the 'target MEI', and the 'target MII'. The participant is first asked to select which image on the right they believe to be the MEI. Next, they are asked to select the image they believe to be the MII. They are given feedback as to the correct choices after every trial. Each participant completes the 15 trials corresponding to those features in a single layer before moving on the next layer. An example trial showing what the participant sees when selecting the MEI and MII is shown in Figure \ref{fig:trial_human_exp}. We recruited 48 participants for this experiment through Prolific (www.prolific.com). They were paid \$3 to complete the \textasciitilde 15 minute experiment.\par
The results of this experiment are in agreement with our hypothesis, and can be seen in Figure \ref{fig:human_exp_accuracy}. Specifically, participants were able to extrapolate from the set of MEIs to the new MEI well across all layers. However, participants could not easily extrapolate from MEIs to their 'opposite' MII. Participants performed significantly better (\(p \approx 1e-5\)) when identifying MIIs in the first layer, Conv20, than the middle and late layers, Mixed4a and Mixed5a.

\section{Dot*Cosine Objective}\label{sec:dot*cosine}
Figure \ref{fig:dot*cosine} shows visually the motivation for the dot*cosine objective used for attribution visualizations. The dot product can optimize for the hidden vector \(\bm{h}\) to simply have a large magnitude, but not really point in the direction \(\bm{v}\). Conversely, cosine similarity can be maximized by inputs with a very low magnitude, which aren't salient to the network. The cosine*dot objective optimizing for \(\bm{h}\) to have a large magnitude and point in the direction \(\bm{v}\). 

\begin{figure}
   \centering
   \includegraphics[width=\textwidth]{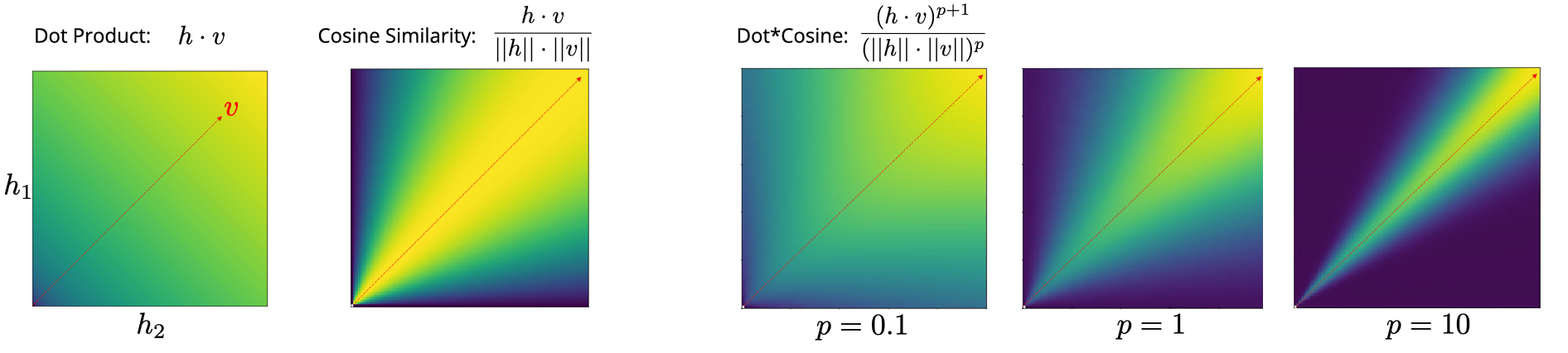}
   \caption{A visual intuition for the dot*cosine loss function}
\end{figure}\label{fig:dot*cosine}

\section{model weight distribution}

\begin{figure}[H]
   \centering
   \includegraphics[width=\textwidth]{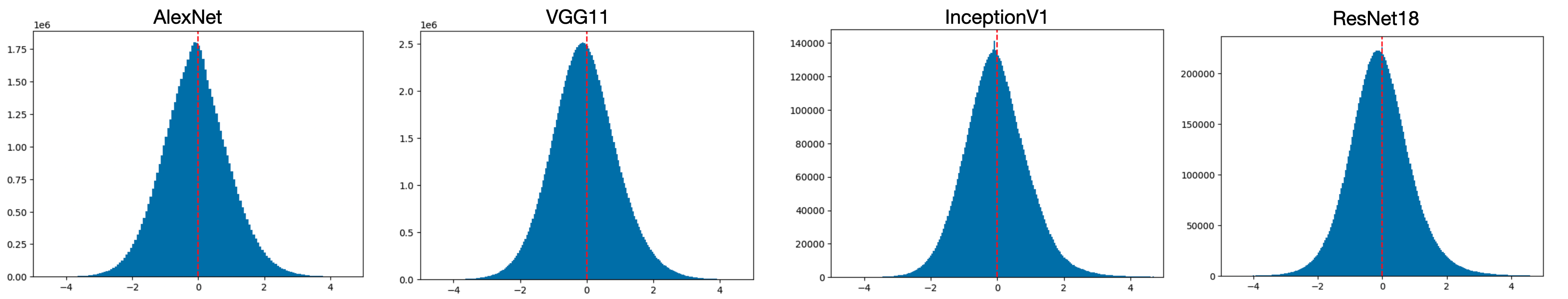}
   \caption{Across Imagenet trained models, we see a very similar weight distribution, with a slight majority of weights being negative in all cases. weights are standardized to \(\sigma = 0\) in each layer before being aggregated in each histogram.}
\end{figure}\label{fig:weight_distribution}

A slight majority of weights are negative consistently across Imagenet trained models, but the function of these weights has received significantly less treatment in the literature.

\section{Other measures of Attribution Completeness}

\begin{figure}[H]
   \centering
   \includegraphics[width=\textwidth]{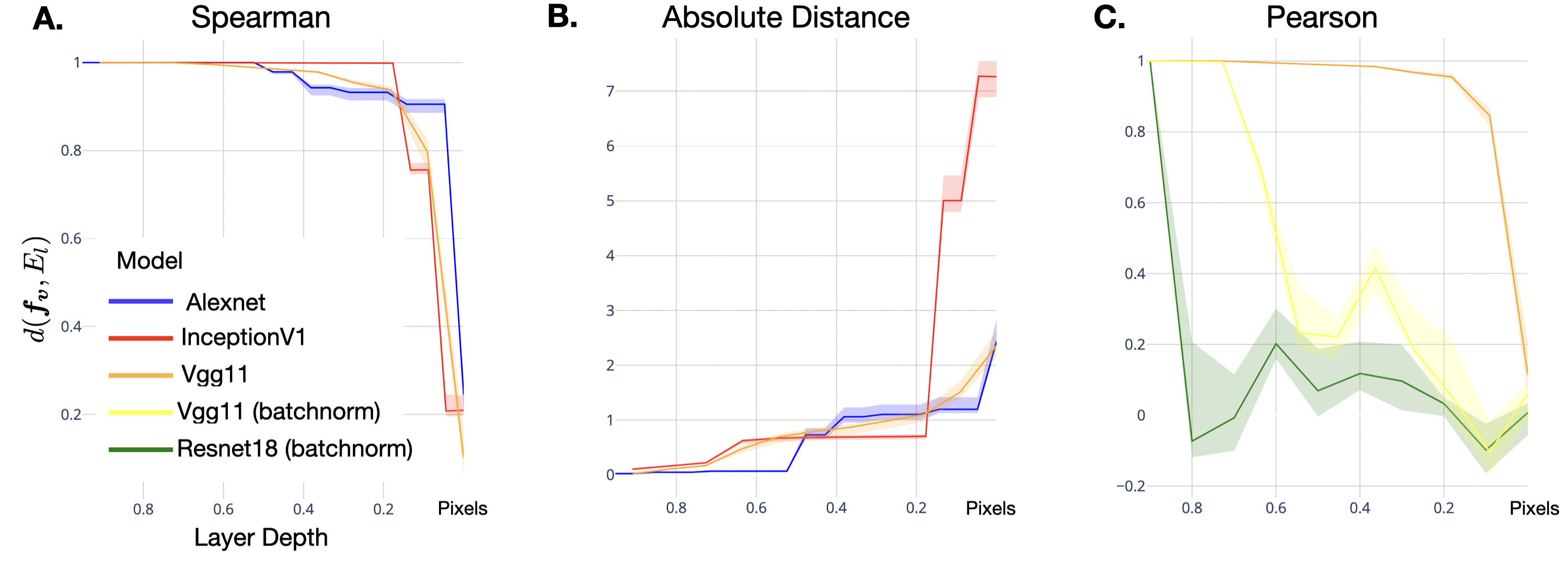}
   \caption{\textbf{A.} The Spearman correlation between logits and total attribution across layers, which is indistinguishable from the Pearson correlation (Figure \ref{fig:energy_correlation}). \textbf{B.} The absolute distance (L1) between the attribution and logit activation. \textbf{C.} The Pearson correlation measured on models with batch normalization layers. VGG11 (no batch normalization) is displayed again in this plot for reference to VGG11 (with batch normalization).}
\end{figure}\label{fig:energy_corr_other}

In section \ref{sec:completeness} we show a ceiling Pearson correlation between model logits and the summed attribution through many preceding layers of the model latent space. Pearson correlation is a good metric as the logit activations are approximately normally distributed, and we dont care about the overall scale of the attributions for our purposes of selecting MTIs. That said, here we report some other distance metrics between the attributions and logit activations; \textbf{A.} Spearman correlations and \textbf{B.} absolute (L1) distance. Additionally in \textbf{C.} we show the original Pearson metric, but measured on models with batch normalization layers. Passing through batch normalization layers causes the correlation between attribution and logit activation to drop significantly. For all these plots, and that in Figure \ref{fig:energy_correlation}, we show a standardized 'model depth' on the X axis in the range \([0,1]\), which corresponds to the ratio of ReLU non-linearities preceding the layer over the total in the model.

\section{Do Attribution Inversions Actually Excite/Inhibit?}

\begin{wrapfigure}{R}{0.45\textwidth}
  \begin{center}
    \includegraphics[width=0.44\textwidth]{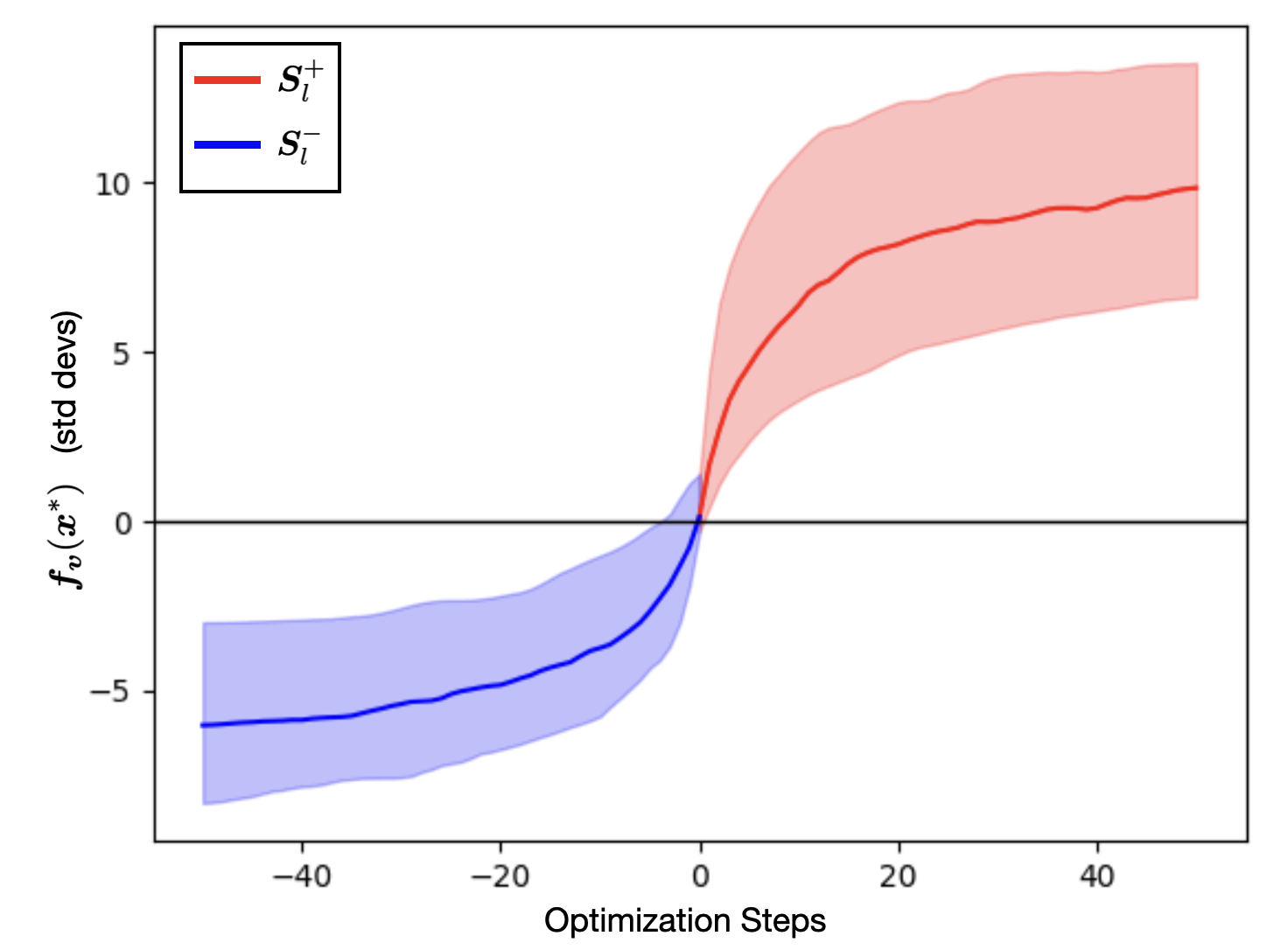}
  \end{center}
    \caption{Activations induced in the target feature by optimizing towards \textcolor{blue}{\(\bm{S}_{l}^{-}\)} and \textcolor{red}{\(\bm{S}_{l}^{+}\)}}\label{fig:attr_act_opt_steps}
\end{wrapfigure}%

Attribution inversion is suppose to tell us something about feature \(\pred_{\vv}\), but optimizes an objective based on the attribution feature vector in an earlier layer \(\bm{S}_{l}^{\pm}\). Here we conduct a simple sanity check to confirm optimizing for \(\bm{S}_{l}^{\pm}\) has the expected effect of exciting/inhibiting \(\pred_{\vv}\). To test this, for 5 layers of InceptionV1 we choose 20 random units, then compute inhibitory and excitatory attribution inversion through the preceding layer across for 10 of the unit's MTIs. Figure \ref{fig:attr_act_opt_steps} shows the activation these inversions induce in the target feature across optimization steps. The shaded region corresponds to the full range of results, and the lines correspond to the median. We plot activation on the y-axis in standard deviations for the target unit. The figure shows that attribution inversions indeed have the desired effect.

\section{Toy model details}

Architecturally our absolute value toy model and xor toy model are nearly identical, differing only in their input dimensions -- 6 for the abs model but 12 the xor model, as each of the 6 output features is a function of a pair of inputs. From these inputs \(\vx\), a hidden vector \(\bm{h}\) is computed, and then the output \(\pred^{\prime}\) as follows;

\begin{align}\label{toy model architecture}
\bm{h} = \text{ReLU}(W_{1}\vx+b_{1}) \\
\pred^{\prime} = \text{ReLU}(W_{2}\bm{h}+b_{2})
\end{align}

These models were trained in a manner similar to that in Elhage et. al. (2022)\cite{superposition}. First, for the absolute value model, \(\vx\) is sampled such that \(x_{i}\) has a \(.99\) probability of being 0 (it is sparse), otherwise it is sampled uniformly from \([0,1]\). Each feature receives an 'importance' \(I_{i} = .9^{i}\), so the loss function can weight important features more heavily. We train the model  using the mean squared error from the target function, \(f_{i} = \text{abs}(x_{i})\);

\begin{equation}\label{eq:toy loss}
\mathcal{L} = \sum_{i=1}^{6}I_{i}(f_{i} - f^{\prime}_{i})^{2}
\end{equation}

\begin{figure}
   \centering
   \includegraphics[width=\textwidth]{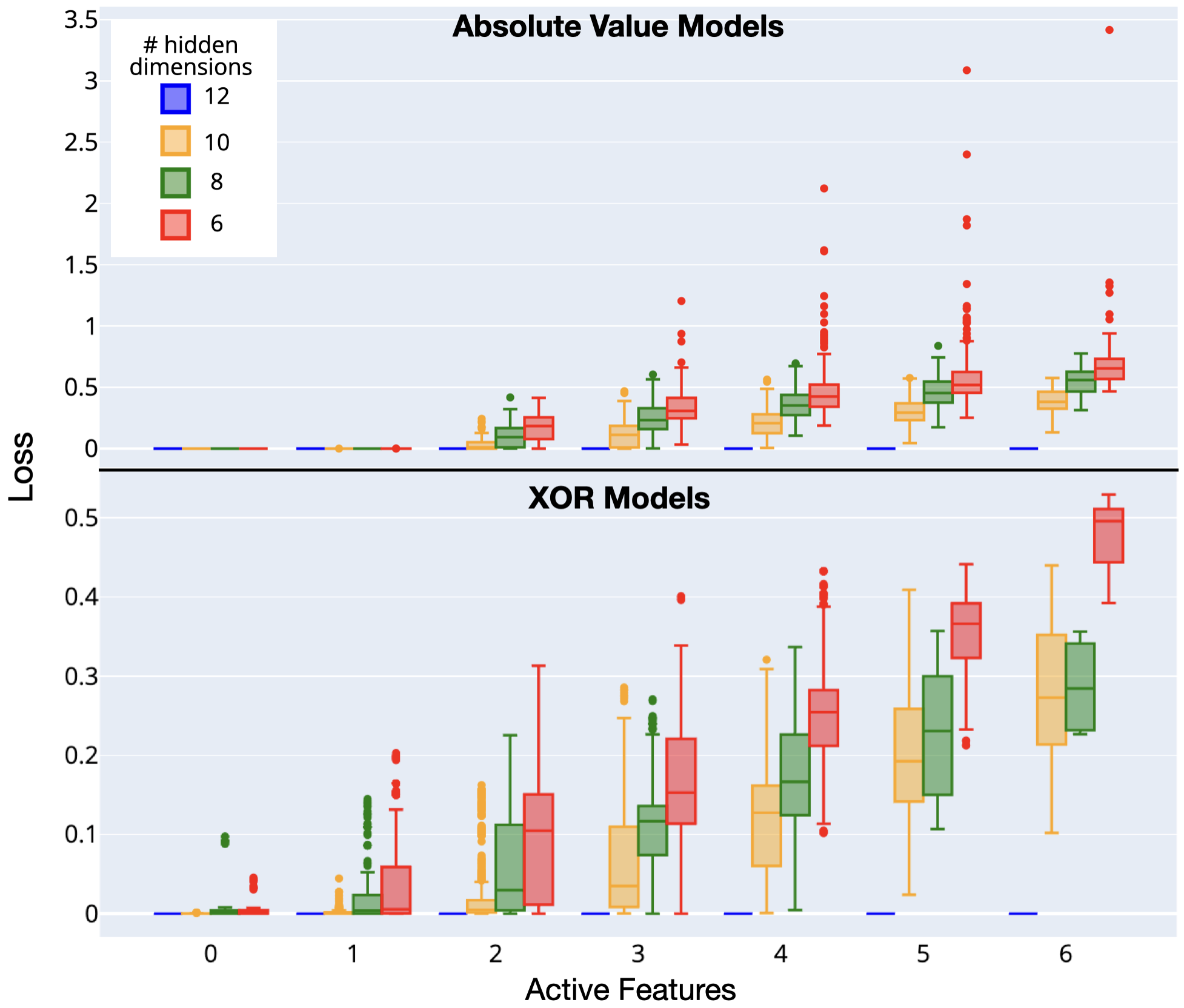}
   \caption{Losses for the Absolute Values and XOR models across inputs with differing number of active features. Where disentangled models \(m=12\) are perfect for all inputs, decreasing the hidden dimensions increases the loss. In particular, loss is greater the more features are present in the input.}
\end{figure}\label{fig:toy model loss}

We train on batches of 600 inputs for 20000 iterations using the Adam optimizer \cite{kingma2017adam}, with learning rate .001. As in the original work we find the model is hard to train on the sparse input signal, and thus train from 1000 different random seeds, picking the top performing model. We repeat this process for the 4 hidden dimensionalities tested; \(m=12,10,8,6\).\par
For the XOR toy model the procedure is largely the same as above, the main difference is how we sample the input. As before, each feature \(f_{i}\) is sampled with some probability \(S\) of being 'off', or 0 in the input. However, in this case when a feature passes this sampling filter and is 'on', it corresponds to a pair of elements in the input, \([x_{2i},x_{2i-1}]\), which are each sampled independently from a Bernoulli distribution, with \(p=.5\). This sampling procedure introduces additional sparsity over the absolute value features, so we use a \(S=.95\) for sampling XOR features. In this case our target function is \(f_{i} = \text{XOR}(x_{2i},x_{2i-1})\), and loss is computed as before (equation \ref{eq:toy loss}. All other training details are identical to the absolute value toy models.\par


It is important we ensure our toy models are actually performant, if we are to take anything from the results in Figure \ref{fig:toy_model_energies}. To test this, for each model, we pass inputs that span the domain and compute the resultant loss. For the absolute values models these inputs are every combination of elements \(\{0,-1,1\}\) in 6 dimensions. For the XOR models, we test every combination of elements \(\{0,1\}\) in 12 dimensions. We show these input-wise losses in Figure \ref{fig:toy model loss} as box plots, organized by the toy model hidden dimensions and the number of 'active features'-- i.e. the number of 1s in the output computed by the ground-truth absolute value and XOR functions. We see the the disentangled models, with 12 hidden units, are perfect, incurring no loss for any of the inputs. As the hidden dimensionality decreases however, both models incur loss as they cannot faithfully represent the target function. Of note, this loss monotonically increases with the number of active features, in agreement with the theory that features in superposition can be faithfully represented when either is present in the input, but not simultaneously present. The inputs used for Figure \ref{fig:toy_model_energies} all have 1 or 0 active features, and the models are performant over these inputs. 

\section{Uniqueness Experiment}\label{sec:appendix uniqueness}

K-means directions in this experiment were defined as the cluster centroids determined by the SKlearn k-means clustering algorithm using the cosine distance metric. The clustering was applied to each layers' hidden vectors in response to the ImageNet validation set, each sampled at a random position in the layer's activation map. We first used k=1000 centroids, to define a general basis of many features. To get our 20 sampled features for the experiment, we ran k-means again on these centroid vectors using k=20, then selected a random vector from the original 1000 from each of these new clusters. We did this to ensure the features were sampled from a reasonably sized basis (larger than 20), but also not too close to each other in direction. Gradients in this experiment were computed with respect to the central position in each activation map.\par

\begin{figure}
   \centering
   \includegraphics[width=\textwidth]{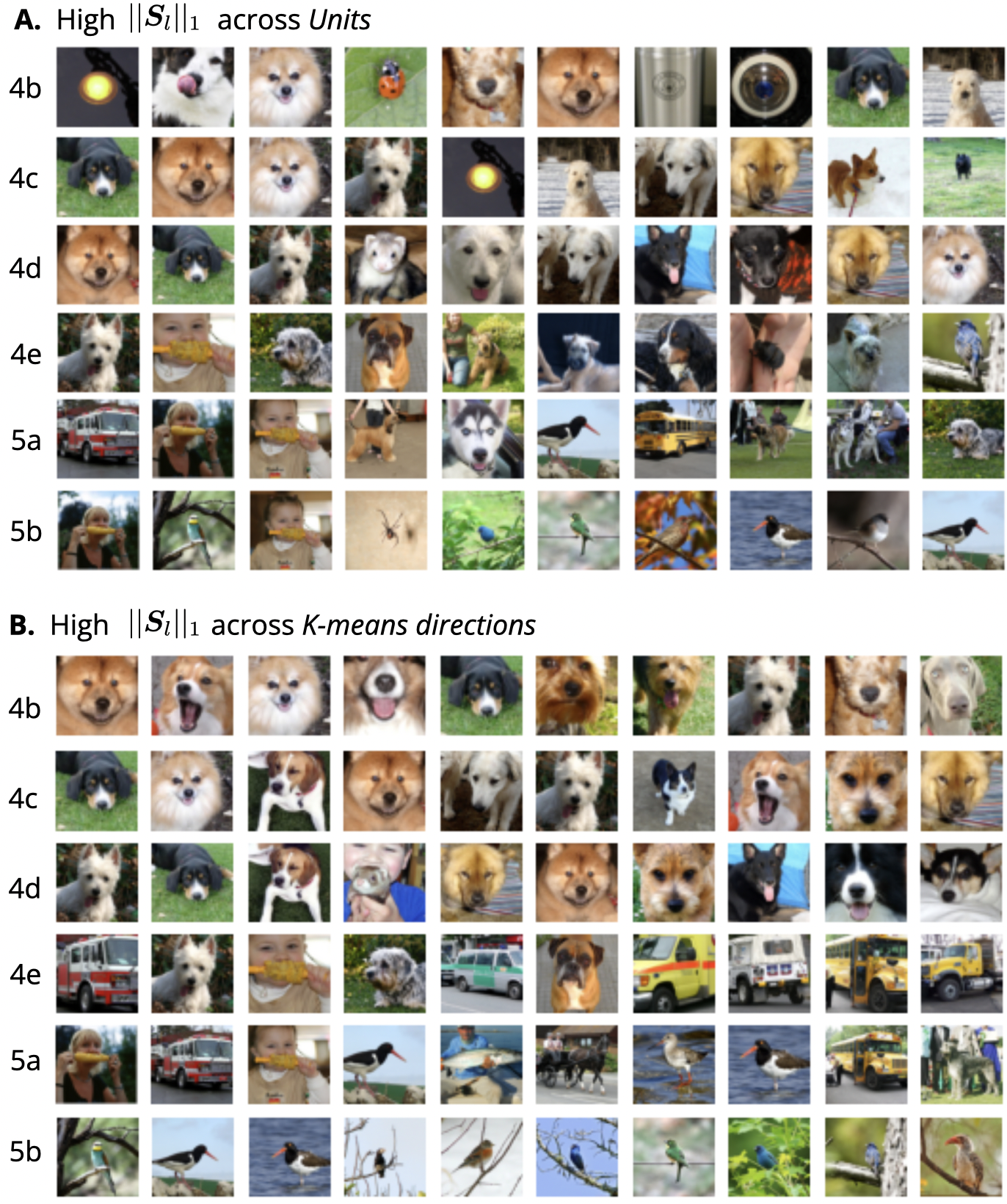}
   \caption{Images with large attributions across \textbf{A.} unit, and \textbf{B.} k-means features.}\label{fig:non-unique}

\end{figure}

We found in this uniqueness experiment that across features the very same images \textit{superposition features} loading onto many units in the attribution layer. Viewing the corresponding images, which are shown in figure \ref{fig:non-unique} helps lend some credence to this view. Each (receptive field cropped) image was among the top-100 largest attributions for at least 15 out of the 20 features sampled. These images show single salient concepts, like 'dog-face', 'bird', or 'emergency vehicle', rather than many concepts. This motivates our hypothesis that these images load onto many units because they express features the model has placed in superposition, rather than the alternative, that they simply express many different features at once. 

\section{More attribution/inversion examples}

\begin{figure}[H]
   \centering
   \includegraphics[width=.99\textwidth]{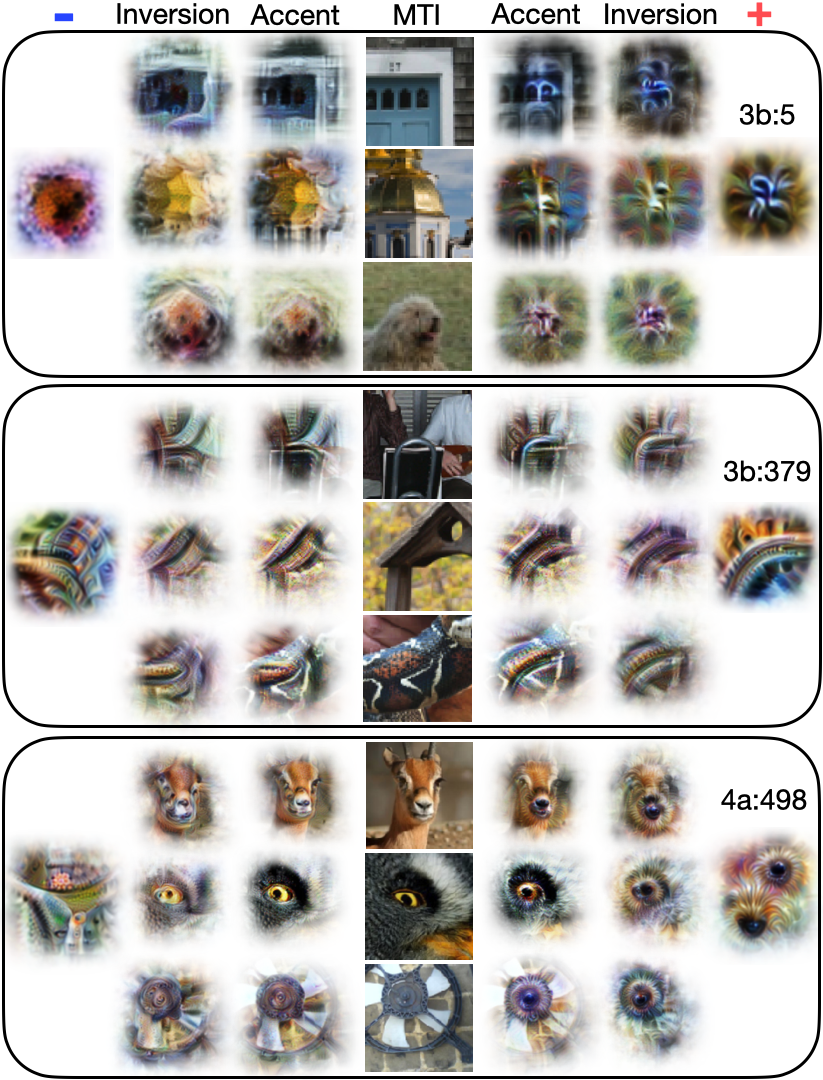}
   \caption{}
\end{figure}\label{fig:MTI_faccent_more}

\begin{figure}[H]
   \centering
   \includegraphics[width=.99\textwidth]{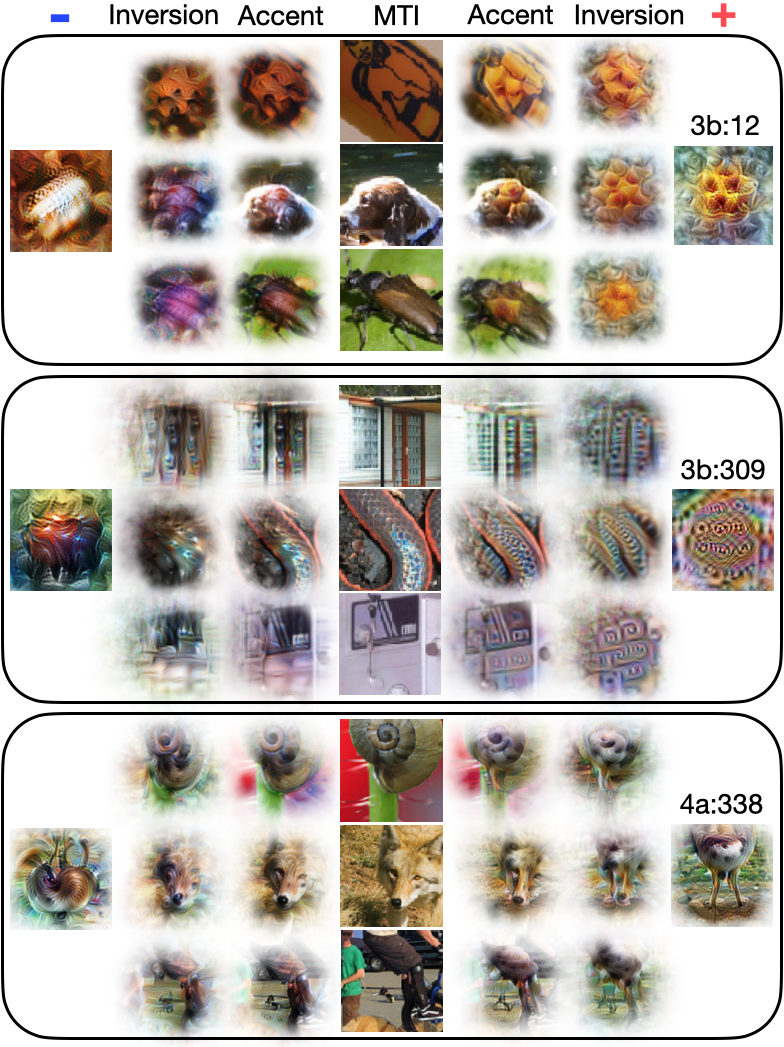}
   \caption{}
\end{figure}\label{fig:MTI_faccent_2}

\section{more atlases}
\begin{figure}[H]
   \centering
   \includegraphics[width=.95\textwidth]{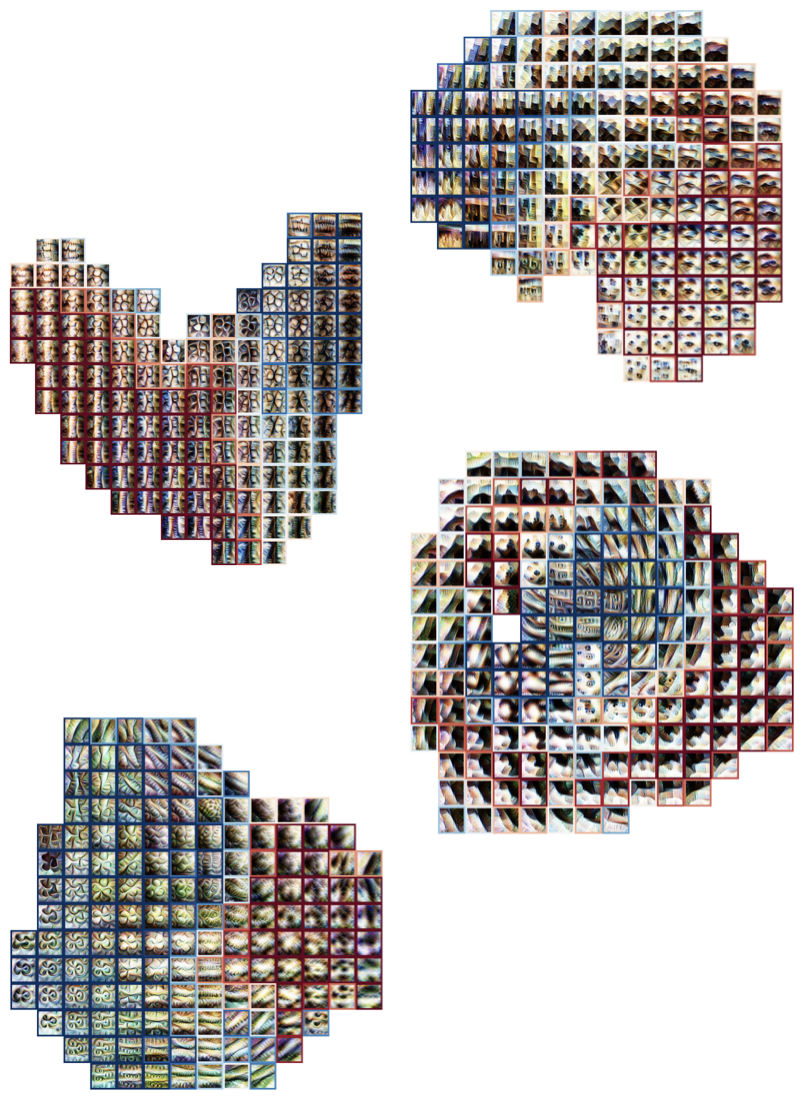}
   \caption{}
   \label{fig:atlases_2}
\end{figure}

\begin{figure}[H]
   \centering
   \includegraphics[width=.95\textwidth]{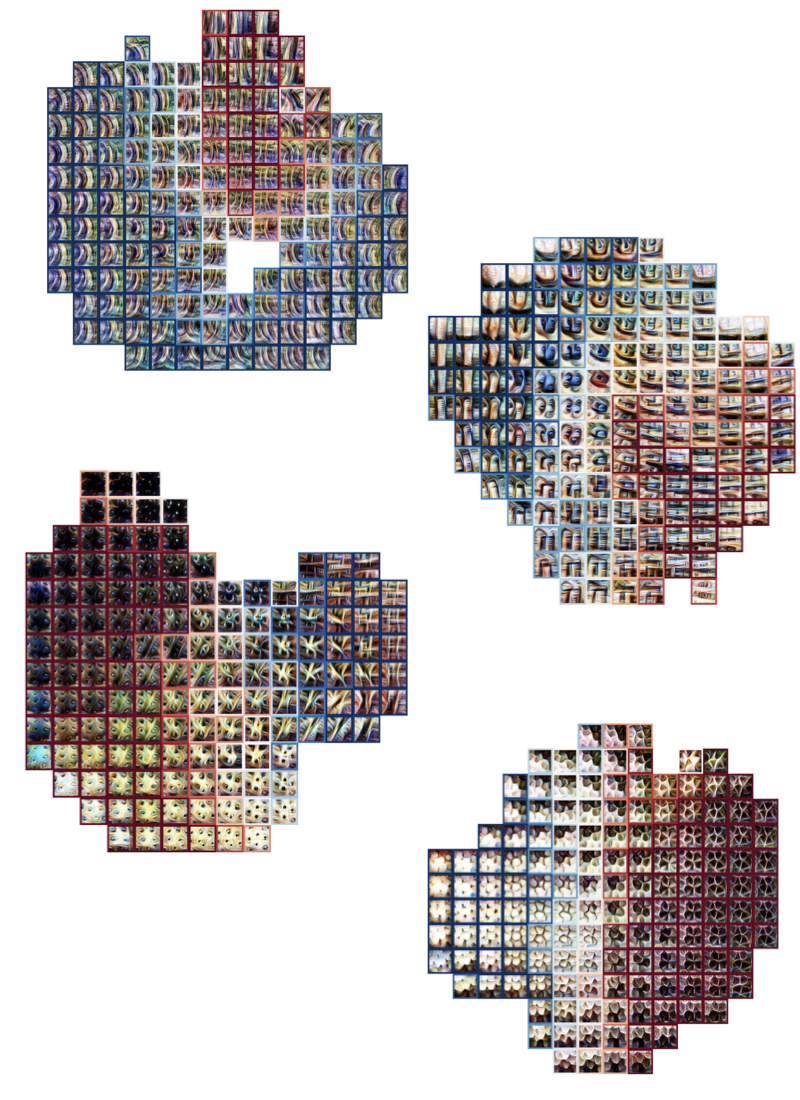}
   \caption{}
   \label{fig:atlases_3}
\end{figure}

\clearpage

\end{document}